%% file: main.tex
\documentclass{article}

\usepackage{microtype}
\usepackage{graphicx}
\usepackage{subcaption}
\usepackage{booktabs} 
\usepackage{arydshln}
\usepackage[table]{xcolor}
\usepackage{makecell}
\usepackage{threeparttable}
\usepackage{caption}
\usepackage[preprint]{icml2026}
\definecolor{SDEblue}{RGB}{28 58 88}
\usepackage{multirow}
\usepackage{amsmath}
\usepackage{amssymb}
\usepackage{mathtools}
\usepackage{amsthm}
\usepackage[colorlinks=true,
            linkcolor=red,
            citecolor=SDEblue,
            urlcolor=blue]{hyperref}

\usepackage[capitalize,noabbrev]{cleveref}

\theoremstyle{plain}

\theoremstyle{definition}

\theoremstyle{remark}

\usepackage[textsize=tiny]{todonotes}

\icmltitlerunning{Rethinking the Role of LLMs in Time Series Forecasting}

\begin{document}

\twocolumn[
  \icmltitle{Rethinking the Role of LLMs in Time Series Forecasting}

  \icmlsetsymbol{equal}{*}

  \begin{icmlauthorlist}
    \icmlauthor{Xin Qiu}{yyy,zju}
    \icmlauthor{Junlong Tong}{yyy} 
    \icmlauthor{Yirong Sun}{yyy}
    \icmlauthor{Yunpu Ma}{lmu}
    \icmlauthor{Wei Zhang}{yyy}
    \icmlauthor{Xiaoyu Shen}{yyy}
  \end{icmlauthorlist}

    \icmlaffiliation{yyy}{Ningbo Key Laboratory of Spatial Intelligence and Digital Derivative, Institute of Digital Twin, Eastern Institute of Technology, Ningbo}  \icmlaffiliation{zju}{Zhejiang University}
    \icmlaffiliation{lmu}{LMU Munich}
    \icmlcorrespondingauthor{Xiaoyu Shen}{xyshen@eitech.edu.cn}

  \icmlkeywords{Machine Learning, ICML}

  \vskip 0.3in
]

\definecolor{green_negative}{HTML}{37B24D}
\definecolor{mygreen}{HTML}{07ad25}
\definecolor{myred}{HTML}{ee1f1f}
\printAffiliationsAndNotice{}
\input{Section/00_abstract}
\input{Section/01_intro}
\input{Section/02_preliminary}

\input{Section/03_cross_domain}
\input{Section/04_Do}
\input{Section/05_need}

\input{Section/06_dis}

\clearpage
\bibliography{example_paper}
\bibliographystyle{icml2026}

\newpage
\appendix
\onecolumn
\input{Section/related_work}

\input{Section/APP_analy}
\input{Section/APP_TSF}
\input{Section/APP_prompt}
\input{Section/APP_baseline}
\input{Section/APP_Dataset}
\input{Section/APP_re}
\input{Section/APP_sy}
\input{Section/APP_router}
\input{Section/APP_repro}
\end{document}

%% file: Section/00_abstract.tex
\begin{abstract}
Large language models (LLMs) have been introduced to time series forecasting (TSF) to incorporate contextual knowledge beyond numerical signals. However, existing studies question whether LLMs provide genuine benefits, often reporting comparable performance without LLMs. We show that such conclusions stem from limited evaluation settings and do not hold at scale.
We conduct a large-scale study of LLM-based TSF (LLM4TSF) across 8 billion observations, 17 forecasting scenarios, 4 horizons, multiple alignment strategies, and both in-domain and out-of-domain settings.
Our results demonstrate that \emph{LLM4TS indeed improves forecasting performance}, with especially large gains in cross-domain generalization. Pre-alignment outperforming post-alignment in over 90\% of tasks. Both pretrained knowledge and model architecture of LLMs contribute and play complementary roles: pretraining is critical under distribution shifts, while architecture excels at modeling complex temporal dynamics. Moreover, under large-scale mixed distributions, a fully intact LLM becomes indispensable, as confirmed by token-level routing analysis and prompt-based improvements.
Overall, our findings overturn prior negative assessments, establish clear conditions under which LLMs are not only useful, and provide practical guidance for effective model design. We release our code at \url{https://github.com/EIT-NLP/LLM4TSF}.
\end{abstract}

%% file: Section/01_intro.tex
\section{Introduction}
Time series forecasting (TSF) is a fundamental learning task with broad applications across many domains~\cite{liu2024deep,kim2025comprehensive,kong2025deep}. Despite decades of research on statistical and machine learning approaches~\cite{de200625,masini2023machine,yunita2025performance}, accurate TSF remains challenging due to the coexistence of long-term trends, periodic patterns, abrupt changes, and stochastic noise. Inspired by the success of Transformers in fields such as speech, vision, and video understanding~\cite{dong2018speech,liu2021swin,liu2022video,cao2022swin}, prior work has introduced Transformer-based architectures to TSF~\cite{wu2021autoformer,zhang2024skip,liang2024crossformer,li2025fcp}. By leveraging attention mechanisms, these models enable more flexible modeling of temporal dependencies. 

However, most Transformer-based TSF models are trained from scratch on uni-modal numerical time series. Such representations are inherently abstract and lack explicit encoding of real-world context or environmental factors that often underlie temporal observations~\cite{liu2024time,huang2025cross,leiitformer}. This limitation has motivated recent explorations of LLM-based time series forecasting (LLM4TSF), aiming to leverage the rich world knowledge embedded in LLMs pretrained on large-scale text corpora~\cite{ICLR2025_e1de63ec,liu2025timecma,hu2025sst,wolff2025using}. To reduce the modality gap between numerical time series and language representations, existing approaches typically follow two alignment paradigms: \emph{Pre-alignment} methods map time series into language-compatible representations via cross-attention with word embeddings before feeding them into an LLM~\cite{ICLR2025_e1de63ec,liu2025timecma}. In contrast, \emph{post-alignment} methods jointly fine-tune time series encoders and LLMs through supervised learning, adapting both components simultaneously~\cite{meunier2025crisists,liu2025calf}.

Despite this progress, a fundamental question remains unresolved: \emph{are LLMs truly indispensable for TSF?} Many recent studies raise doubts, arguing that existing alignment strategies may induce only pseudo-alignment~\cite{zheng2025understanding,zheng2025liftingmanifoldsmitigatepseudoalignment}, or observing that removing the LLM module causes little to no performance degradation~\cite{tan2024language,zheng2025understanding,zhang2025text}. These findings have sparked an ongoing debate about whether LLMs contribute genuine modeling capability, or merely act as architectural or parameter-level augmentations.

We argue that existing analyses are insufficient to answer this question conclusively. Prior studies are typically conducted on small-scale datasets, rely on only the shallow layers of LLMs~\cite{tan2024language,zheng2025understanding}, focus primarily on in-domain evaluation~\cite{zhang2025text}, and rarely probe the underlying mechanisms responsible for performance differences. Importantly, the core strength of LLMs lies not in their architecture alone, but in their pretrained world knowledge, instruction-following ability, and capacity for multi-task generalization~\cite{brown2020language}. Evaluating LLM4TSF under single-task, in-domain settings with partially utilized LLM parameters fails to reflect these capabilities and may lead to misleading conclusions.

To address these limitations, we conduct a large-scale, systematic study of LLM4TSF across diverse settings. Our evaluation spans 8 billion observations, 17 forecasting scenarios, both in-domain and out-of-domain distributions, and four forecasting horizons. We examine representative pre-alignment and post-alignment strategies, and explicitly disentangle the roles of pretrained knowledge, model architecture, and alignment design in LLM4TSF.

Our empirical analysis provides clear evidence that \emph{LLM4TSF indeed improves forecasting performance}. First, alignment strategy plays a decisive role: \emph{pre-alignment methods outperform post-alignment} approaches in over 90\% of tasks overall. Second, performance gains arise from \emph{a complementary interaction between pretrained knowledge and architectural capacity}. Pretraining proves particularly valuable under distribution shifts and out-of-domain settings, while architectural components excel at capturing complex temporal dynamics. Third, \emph{data diversity is critical}: models trained on multi-source time series consistently outperform single-dataset baselines in more than 70\% of in-domain tasks and exhibit stronger cross-domain generalization than TS-specific models. Finally, \emph{LLM4TSF models show clear preferences for certain statistical regimes}, performing especially well on data with frequent transitions and high variability.
Further analysis under mixed-distribution and large-scale settings yields additional insights. Unlike low-data regimes where randomizing or removing the LLM has minimal impact~\cite{tan2024language}, we find that a fully intact LLM becomes essential for robust forecasting at scale, and partial fine-tuning is no longer sufficient. Token-level routing analysis provides mechanistic evidence for this effect: the model’s decision to route tokens through or around the LLM strongly correlates with forecasting errors, indicating adaptive utilization of LLM capabilities. Moreover, informative textual prompts consistently improve performance, underscoring the importance of semantic guidance beyond simply increasing model size.

At the same time, our study highlights clear limitations. LLM4TSF does not automatically benefit from larger LLMs without careful alignment, and performance remains sensitive to data distribution, preventing uniformly strong results across all scenarios.
In summary, our main contributions are as follows:
\begin{enumerate}
    \item \emph{A large-scale empirical study} that provides the first comprehensive assessment of the benefits and limitations of LLM4TSF.
    \item \emph{A principled decomposition of performance gains}, clarifying the distinct and complementary benefits of pretrained knowledge and model architecture.
    \item \emph{A routing-based analysis} that links token-level path selection to macroscopic forecasting performance, offering concrete evidence of LLM effectiveness.
    \item \emph{Practical guidelines and capability boundaries for applying LLMs to TSF}, informing the design of future LLM-based forecasting systems.
\end{enumerate}

%% file: Section/02_preliminary.tex
\section{Preliminary}
\subsection{Single \& Cross-Dataset Learning Paradigm}
\label{Single-Dataset & Cross-Dataset Learning in TSF}
TSF applications often involves TS data from diverse domains
with substantial differences in statistical properties~\cite{liu2024unitime,chang2025llm4ts}.
Ideally, a model should not only achieve strong performance on a single dataset,
but also be capable of transferring knowledge across heterogeneous datasets~\cite{cheng2024disentangled,xiao2025timefoundfoundationmodeltime}.
However, many existing LLM4TSF still adopt a \emph{\textbf{single-dataset learning}} paradigm.
For example, S$^2$IP~\cite{pan2024s}, FSCA~\cite{ICLR2025_e1de63ec}, TransDF~\cite{wang2025transdf},  and CALF~\cite{liu2025calf} are typically trained and evaluated on individual datasets.
Such settings are prone to overfitting on limited data
and limit the potential generalization advantages of LLMs.
Inspired by domain-specific TS foundation models trained from scratch on large-scale data~\cite{ansari2024chronos,ning2025tsrag,liu2025sundialfamilyhighlycapable}, prior work such as UniTime~\cite{liu2024unitime} adopts \emph{\textbf{cross-dataset learning}} for LLM4TSF, enabling stable in-domain and out-of-domain generalization, while we further introduce fine-grained instructions to support instruction-driven task generalization~\cite{zhou2023instruction}.

\subsection{Core components of LLM4TSF}
\label{architecture}
Typical architecture of LLM4TSF models consist of three core components: \emph{\textbf{TS encoder}}, \emph{\textbf{LLM backbone}} and \emph{\textbf{TS decoder}}. Both the TS encoder and decoder are implemented as lightweight MLPs,
decoupling low-level numerical processing from high-level learning~\cite{chencloser}.
The LLM backbone is instantiated with pre-trained LLM.

\emph{\textbf{(I) TS encoder.}} Given a TS $\mathbf{X}_{1:L} \in \mathbb{R}^{L \times d}$, channel-independent and instance normalization strategies are applied to mitigate scale variations across variables~\cite{kim2021reversible}.
Subsequently, a patching operation is employed to divide each  TS into a sequence of local patches~\cite{,Yuqietal-2023-PatchTST}.
Specifically, we denote the patch length as $P$ and the stride as $S$.
The patching produces patch-level 
$\mathbf{x}_{p} \in \mathbb{R}^{P \times N}$,
where $N$ denotes the patch numbers, defined as
$
N = \frac{L - P}{S} + 2 .
$
The TS embeddings are then obtained as
$
\mathbf{X} = f_{\mathrm{enc}}\!\left(\mathbf{x}_{p}\right),
$
where $f_{\mathrm{enc}}(\cdot)$ denotes the TS encoder.
\emph{\textbf{(II) LLM backbone.}}
To activate the prior knowledge in the LLM, textual prompts are introduced to describe background information and task specifications.
The text prompts are processed by the LLM tokenizer to obtain prompt embeddings.
At the input layer of the LLM, the prompt embeddings are fed with the TS embeddings.
The resulting hidden states are used as the integrated representations, expressed as
$
\mathbf{h} =
f_{\mathrm{LLM}}\!\left(\mathbf{Z}_{\mathrm{}
, }\mathbf{X}
\right),
$
where $\mathbf{Z}_{\mathrm{}}$ and $\mathbf{X}$ denote the prompt and TS embeddings.
\emph{\textbf{(III) TS decoder.}} A TS decoder maps the  outputs $\mathbf{h}$ back to forecasts:
$
\hat{\mathbf{X}}_{L+1:L+H} = f_{\mathrm{dec}}\!\left(\mathbf{h}\right),
$
where $f_{\mathrm{dec}}(\cdot)$ denotes the TS decoder.

\subsection{Alignment Strategy for LLMs on TSF}
\label{strategy}
\begin{figure}[h]
    \centering
    \includegraphics[width=1\linewidth]{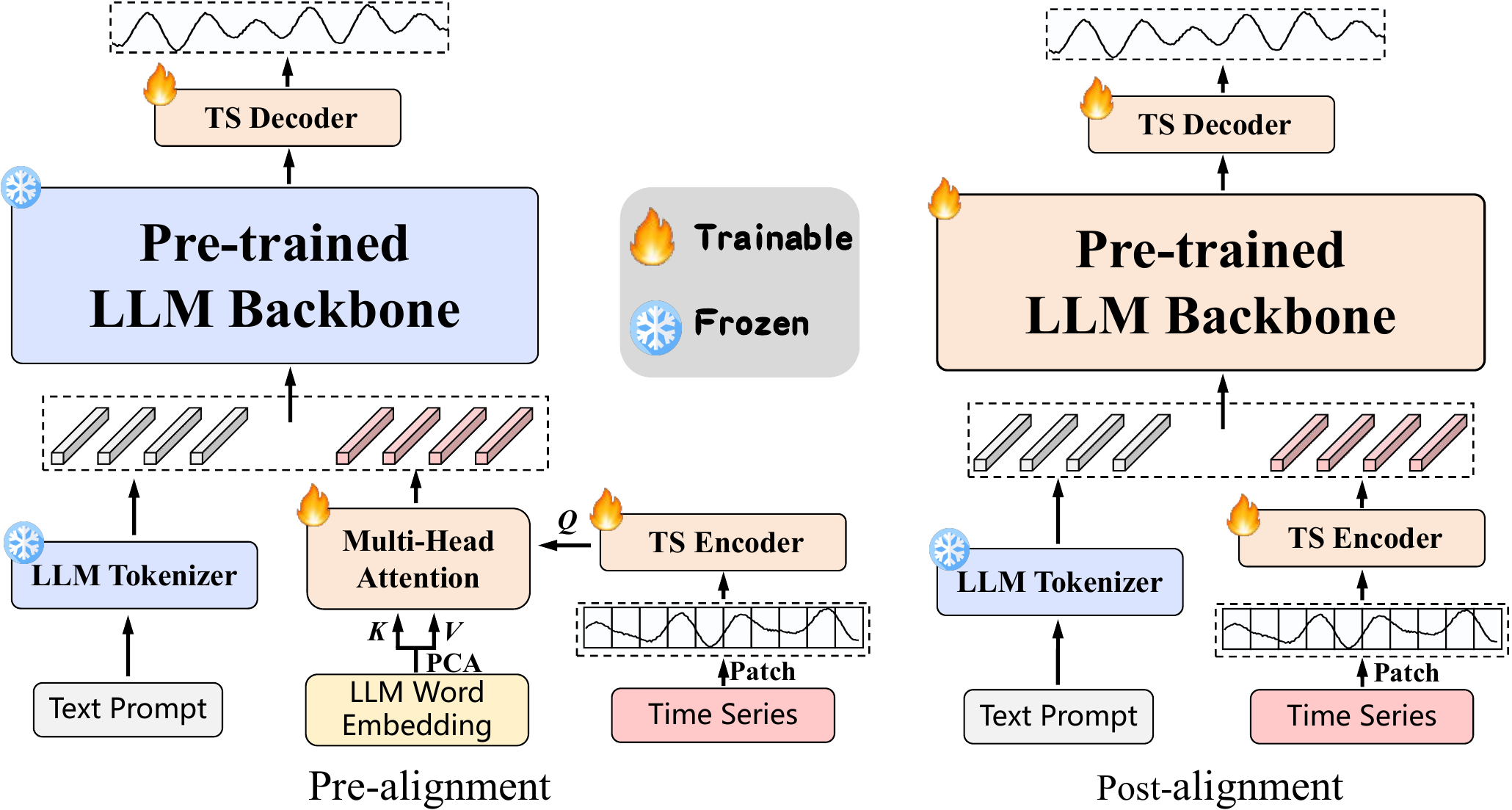}
    \caption{Two mainstream alignment strategies for LLM4TSF.}
    \label{fig:alignment}
\end{figure}

Applying LLMs to TSF involves bridging the modality gap between TS and text modalities.
Existing methods mainly adopt one of two strategies to enable effective cross-modal interaction~\cite{jin2023time,woo2024unified,liu2025calf,meunier2025crisists,ICLR2025_e1de63ec}, namely \emph{\textbf{pre-alignment}} and \emph{\textbf{post-alignment}}, as illustrated in Fig.~\ref{fig:alignment}.

\textbf{Pre-alignment.}\quad
Pre-alignment aligns TS to the textual modality \emph{before} input to the LLM,
exploiting the semantic structure of pre-trained word embeddings while keeping the LLM frozen.
Let $\mathbf{X}\in \mathbb{R}^{N \times M}$ denote the TS embeddings to be aligned,
where $N$ is the number of TS tokens and $M$ is the embedding dimension.
Let $\mathbf{D} \in \mathbb{R}^{|\mathcal{A}| \times M}$ denote the word embedding dictionary of the LLM,
where $|\mathcal{A}|$ is the vocabulary size.
Due to the large size of $\mathbf{D}$, directly aligning TS embeddings with the full dictionary is expensive.
Therefore, principal component analysis (PCA) is applied to obtain a set of principal word embeddings:
$
\hat{\mathbf{D}} = \mathrm{PCA}(\mathbf{D}),
$
where $\hat{\mathbf{D}} \in \mathbb{R}^{d \times M}$ and $d \ll |\mathcal{A}|$.
Alignment is performed via attention, using TS embeddings as queries and the principal word embeddings as keys and values.

\textbf{Post-alignment.}\quad
Post-alignment perform modality alignment between TS and text within the representation space of the LLM,
by jointly modeling TS embeddings and textual embeddings.
In this paradigm, TS embeddings 
and prompt embeddings 
are fed into the LLM for cross-modal modeling.
Let $\mathbf{X} \in \mathbb{R}^{N \times M}$ denote the TS embeddings
and $\mathbf{Z} \in \mathbb{R}^{C \times M}$ denote the prompt embeddings, where $C$ is the number of text tokens.
The LLM produces integrated representations as:
$
\mathbf{h} =
f_{\mathrm{LLM}}\!\left(
\mathbf{Z}, \mathbf{X}
\right),
$
where $f_{\mathrm{LLM}}(\cdot)$ denotes the forward mapping of the LLM.
During training, the parameters of the LLM are updated using supervision from the TSF task,
thereby enabling alignment between TS and text modalities in the latent space.

%% file: Section/03_cross_domain.tex
\section{Benefits of Diverse TS Data in LLM4TSF}
\label{Exploring the Generalization of LLMs for TSF}
Although prior studies have examined LLM-based TSF, their evaluations are mostly conducted under single distribution. In addition, many approaches adapt LLMs using only shallow layers~\cite{pan2024s,liu2025timecma}, restricting the capacity of pretrained models and making results sensitive to overfitting or dataset-specific artifacts. Consequently, the true capability of LLMs in TSF remains difficult to assess~\cite{tan2024language,zheng2025understanding}.
More importantly, pretrained LLMs are designed to learn transferable representations from diverse data. When evaluated on a single TS dataset, this potential may be underutilized. Motivated by this observation, we adopt cross-dataset learning with full-scale LLMs and compare its performance against single-dataset learning under both in-domain and out-of-domain settings, referring to Appendix~\ref{app_Time Series Forecasting Task}.
\subsection{Experimental Setup}\label{Experimental Setup}
\textbf{Datasets.}\quad
We conduct experiments on a collection of $62$ real-world, publicly available TS datasets spanning over $10$ application domains.
The entire dataset collection is denoted as $\mathcal{D}$ and is partitioned into two disjoint subsets,
$\mathcal{D} = \mathcal{D}_{A} \cup \mathcal{D}_{B}$.
The subset $\mathcal{D}_{A}$ contains $55$ datasets and is used for model development,
while $\mathcal{D}_{B}$ consists of the remaining $7$ datasets, which are completely excluded from training and used solely for out-of-domain evaluation.
For each dataset $\mathcal{D}_i \in \mathcal{D}_{A}$, we apply train--test splits,
where the training split is used for model optimization and the held-out test split is used to evaluate in-domain performance.
Overall, the combined datasets comprise over $8$B observations, providing a diverse testbed for studying in-domain and out-of-domain setting, referring to Appendix~\ref{app_General Overview}.

\textbf{Models.}\quad
We consider both the \emph{\textbf{pre-alignment}} and \emph{\textbf{post-alignment}} strategies introduced in Sec.~\ref{strategy}.
Under each alignment strategy, models are trained following two learning paradigms described in
Sec.~\ref{Single-Dataset & Cross-Dataset Learning in TSF}, namely
\emph{\textbf{single-dataset learning}} and \emph{\textbf{cross-dataset learning}}.
This results in two model variants, denoted as
\emph{\textbf{LLM4TSF (Pre-align)}} and \emph{\textbf{LLM4TSF (Post-align)}}.
The core components of all models follow the architecture described in Sec.~\ref{architecture}.
Specifically, we adopt GPT-2~\cite{radford2019language} as the LLM backbone.
To preserve the full modeling capacity of the pretrained LLM, no layer truncation is applied.
At the input stage, for the TS component, the look-back window length is set to $T=512$.
Non-overlapping patch-level sampling is applied with patch size $P=32$ and stride $S=32$,
and each TS is processed using channel-independent strategy and RevIN normalization.
For the text prompts component, dataset identifiers, background information, and statistical descriptors associated with each TS instance are provided as inputs.
All statistical descriptors are computed solely from the $512$-step look-back window, ensuring that no future information is leaked, referring to Appendix~\ref{app_Text Prompt}. 

\textbf{Test Details.}\quad
The forecasting horizon $H$ is evaluated at $\{96, 192, 336, 720\}$, covering short-term and long-term forecasting scenarios.
Generalization is assessed under two settings: \emph{\textbf{in-domain}} and \emph{\textbf{out-of-domain test}}.
The former is conducted on $10$ datasets with held-out test splits,
and the latter is performed on $7$ datasets that are completely excluded from training, referring to Appendix~\ref{app_In-domain and Cross-domain Evaluation}.
Model performance is measured using MAE and MSE.

\begin{figure}[h]
    \centering
    \includegraphics[width=0.99\linewidth]{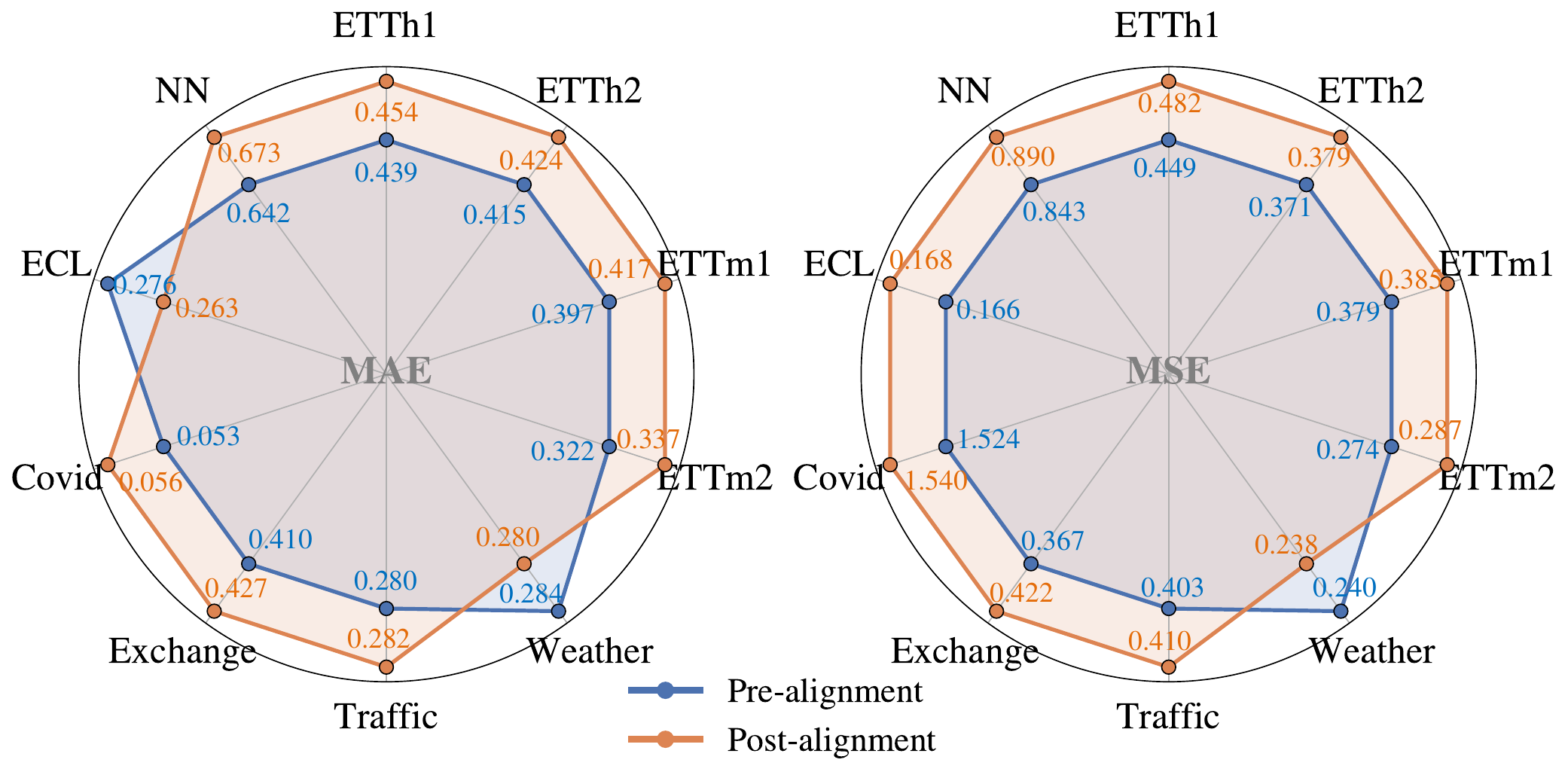}
    \caption{Comparison of LLM4TSF with pre-alignment and post-alignment strategies under single-dataset learning.}
    \label{fig:mae_mse_polar}
\end{figure}

\begin{figure*}[t]
\centering
\includegraphics[width=\textwidth]{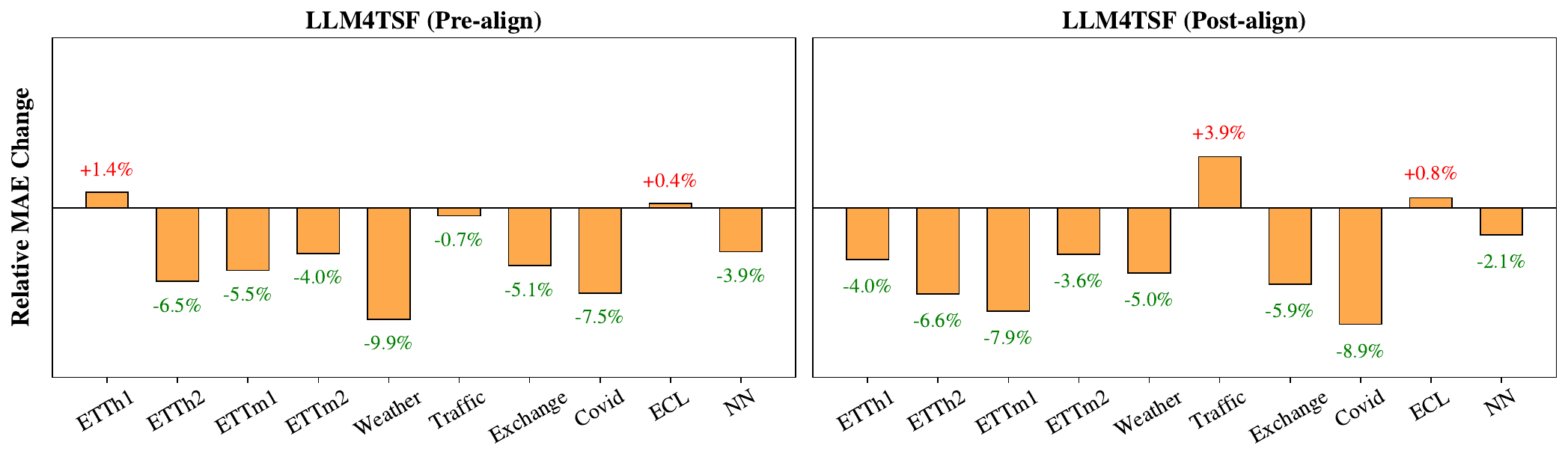}
\caption{Comparison of LLM4TSF performance with pre- and post-alignment under single- and cross-dataset paradigm. 
{\color{green_negative}{Negative}} and \textcolor{red}{Positive} values indicate MAE decreases and increases under cross-dataset learning compared to single-dataset learning.}
\label{fig:llm4tsf_mae}
\end{figure*}

\subsection{Evaluation Results}
\textbf{In-domain Test.}\quad
First, we evaluate the in-domain performance of LLM4TSF under both alignment strategies when trained independently on individual datasets, reporting results across 10 benchmark datasets.
Next, we conduct large-scale cross-dataset joint training and compare the resulting performance gains relative to single-dataset learning.
The forecasting horizon $H$ is evaluated at $\{96, 192, 336, 720\}$; due to space constraints, results are reported as the average over the four horizons. As shown in Fig.~\ref{fig:mae_mse_polar}, under the single-dataset learning paradigm, LLM4TSF with the pre-alignment strategy outperforms its post-alignment counterpart across most datasets.
The performance gap is particularly pronounced on small-scale datasets that are prone to overfitting, such as ETT and Exchange, where pre-alignment demonstrates clear advantages. In addition, we observe that cross-dataset learning leads to better performance, regardless of whether pre-alignment or post-alignment strategies are adopted, as shown in Fig.~\ref{fig:llm4tsf_mae}.
This observation suggests that large-scale training on diverse TS data is more effective, and full results can be found in Appendix~\ref{app_Single- and Cross-Dataset Evaluation Results}.

\begin{table}[t]
\centering
\caption{Out-of-domain test performance (MSE, averaged over horizons $\{96, 192, 336, 720\}$). \textbf{Bold} and \underline{underlined} indicate the best and second-best results for each dataset.}
{\renewcommand{\arraystretch}{1.4}
\resizebox{1.0\linewidth}{!}{
\begin{tabular}{c|cc|ccc|cc}
\toprule
\textbf{Types} & \multicolumn{5}{c|}{\textbf{Zero-Shot Test}} & \multicolumn{2}{c}{\textbf{5\% Few-Shot Test}}\\
\midrule
\textbf{Models} 
& \textbf{Pre-align} & \textbf{Post-align} 
& \textbf{Chronos} 
& \textbf{UniTS} 
& \textbf{Moirai} 
& \textbf{UniTime} 
& \textbf{TimeLLM} \\
\midrule
Wind     & \underline{1.015} & \textbf{0.963} & 1.422 & 1.358 & 1.236 & 1.358 & 1.321 \\
Solar    & \underline{0.228} & 0.274 & 0.434 & 0.871 & 0.936 & \textbf{0.218} & 0.577 \\
AQShunyi & \textbf{0.612} & 0.688 & 0.808 & 0.890 & \underline{0.668} & 0.905 & 0.859 \\
CzenLan  & \textbf{0.274} & \underline{0.288} & 0.298 & 0.738 & 0.660 & 0.401 & 0.319 \\
ZafNoo   & \underline{0.547} & 0.583 & 0.550 & 0.668 & \textbf{0.543} & 0.803 & 0.594 \\
NASDAQ   & \textbf{0.735} & \underline{0.749} & 0.873 & 1.120 & 1.067 & 1.122 & 0.983 \\
PEMS     & \underline{0.256} & 0.291 & 0.686 & 1.303& \textbf{0.243} & 0.419 & 0.416 \\
\bottomrule
\end{tabular}}}
\label{tab:out-of-domain}
\end{table}
\textbf{Out-of-domain Test.}\quad
To evaluate the out-of-domain performance of the two alignment strategies after cross-dataset learning,
we compare LLM4TSF (Pre-align) and LLM4TSF (Post-align) on seven datasets against three large-scale TS foundation models trained from scratch, namely Chronos~\cite{ansari2024chronos}, UniTS~\cite{gao2024units}, and Moirai~\cite{moirai}.
All comparisons are conducted under a zero-shot setting, and the model configurations and parameters are taken directly from the original papers.
In addition, we include two LLM-based TSF models, UniTime~\cite{liu2024unitime} and TimeLLM~\cite{jin2023time}, which are trained using single-dataset few-shot learning with only 5\% of the training data, for further comparison.
As shown in Tab.~\ref{tab:out-of-domain}, both {LLM4TSF (Pre-align)} and {LLM4TSF (Post-align)}
achieve overall superior performance in zero-shot test compared to TS foundation models.
Moreover, they even outperform LLM-based TSF models trained with 5\% data. Full results are reported in Appendix~\ref{app_Baseline Results in Out-of-Domain Tests}.

\textbf{\emph{These results indicate that, when trained with pretrained LLM parameters and large-scale TS data,
not only performs well in in-domain settings
but also exhibits strong cross-domain generalization. Moreover, this advantage is amplified with increasing data scale, as shown in Fig.~\ref{fig:data_ratio_mae_mse}.}}

%% file: Section/04_Do.tex
\section{Where do the Gains Really Come from?}
\label{Do the Gains Really Come from LLM Parameters?}
Sec.~\ref{Exploring the Generalization of LLMs for TSF} shows that LLM4TSF trained under cross-dataset learning achieves strong performance across a wide range of scenarios. However, it remains unclear where these performance gains actually come from. In this section, we conduct a attribution study by either removing the LLM module or randomly initializing parameters, thereby disrupting the pretrained knowledge and keeping the overall architecture and settings unchanged. This design enables us to isolate and rigorously evaluate the actual contribution.
\begin{figure}[h]
    \vspace{-0.3cm}
    \centering
    \includegraphics[width=0.99\linewidth]{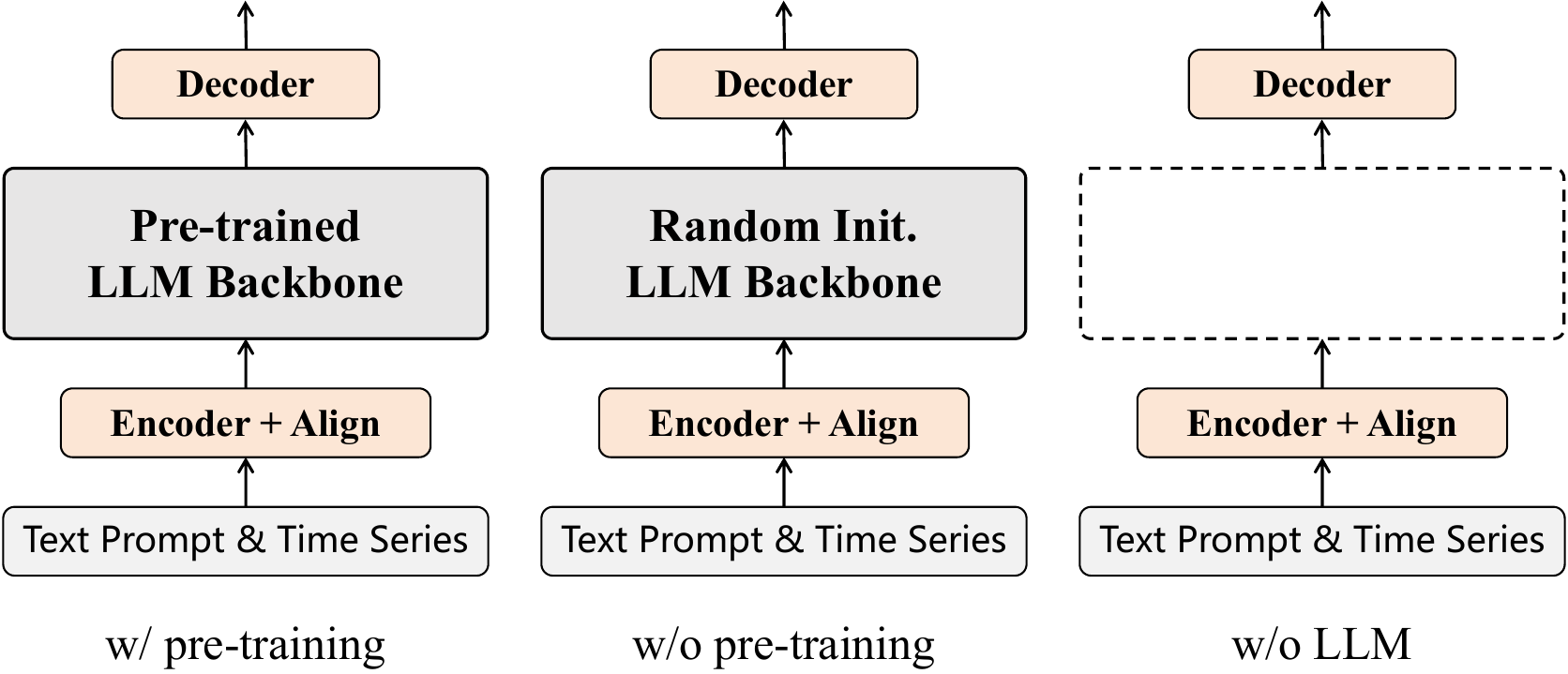}
    \caption{Original architecture and two ablation variants.}
    \label{fig:wo}
    \vspace{-0.5cm}
\end{figure}

\begin{table*}[t]
\centering
\caption{Ablation results of \textbf{LLM4TSF (Pre-align)} and \textbf{LLM4TSF (Post-align)}, averaged over horizons $\{96, 192, 336, 720\}$. For each dataset, the best MAE is highlighted in \textcolor{red}{red}, and the best MSE is highlighted in \textcolor{blue}{blue}, respectively.}

{\renewcommand{\arraystretch}{1.00}
\resizebox{0.98\linewidth}{!}{
\begin{tabular}{c||cc|cc|cc||cc|cc|cc}
\toprule
\multirow{3}{*}{\textbf{Models}} &\multicolumn{6}{c||}{\textbf{LLM4TSF (Pre-align)}} & \multicolumn{6}{c}{\textbf{LLM4TSF (Post-align)}}\\
\cmidrule(lr){2-7} \cmidrule(lr){8-13}
&\multicolumn{2}{c|}{\textbf{w/ Pre-training}}& \multicolumn{2}{c|}{\textbf{w/o Pre-training}}& \multicolumn{2}{c||}{\textbf{w/o LLM}} &\multicolumn{2}{c|}{\textbf{w/ Pre-training}}& \multicolumn{2}{c|}{\textbf{w/o Pre-training}}& \multicolumn{2}{c}{\textbf{w/o LLM}}\\
 & \textbf{MAE} & \textbf{MSE}& \textbf{MAE} & \textbf{MSE}& \textbf{MAE} & \textbf{MSE} & \textbf{MAE} & \textbf{MSE}& \textbf{MAE} & \textbf{MSE}& \textbf{MAE} & \textbf{MSE}\\
\midrule 
\rowcolor{gray!35}\multicolumn{13}{c}{\textbf{In-Domain Test}} \\
\midrule
ETTh1 
& 0.445 & 0.447 
& 0.460 & 0.471 
& \textcolor{red}{0.432} & \textcolor{blue}{0.438} & 0.436 & 0.431
& \textcolor{red}{0.428} & \textcolor{blue}{0.420}
& 0.435 & 0.442\\
ETTh2 
& \textcolor{red}{0.388} & \textcolor{blue}{0.348} 
& 0.414 & 0.377 
& 0.407 & 0.363 & \textcolor{red}{0.396} & \textcolor{blue}{0.357}
& 0.429 & 0.403
& 0.439 & 0.424\\
ETTm1
& 0.375 & 0.353 
& 0.393 & 0.382 
& \textcolor{red}{0.363} & \textcolor{blue}{0.328} & 0.384 & 0.354
& \textcolor{red}{0.373} & \textcolor{blue}{0.340}
& 0.378 & 0.351\\
ETTm2
& \textcolor{red}{0.309} & \textcolor{blue}{0.252} 
& 0.330 & 0.290 
& 0.315 & 0.271 & \textcolor{red}{0.325} & \textcolor{blue}{0.265}
& 0.341 & 0.290
& 0.355 & 0.307 \\
Weather
& \textcolor{red}{0.256} & \textcolor{blue}{0.225} 
& 0.275 & 0.244 
& 0.268 & 0.240 & \textcolor{red}{0.266} & \textcolor{blue}{0.226}
& 0.275 & 0.242
& 0.287 & 0.261\\
Traffic
& 0.278 & 0.401 
& 0.285 & 0.416 
& \textcolor{red}{0.263} & \textcolor{blue}{0.377} & 0.293 & 0.418
& 0.288 & 0.406
& \textcolor{red}{0.281} & \textcolor{blue}{0.390}\\
Exchange
& \textcolor{red}{0.389} & \textcolor{blue}{0.332} 
& 0.423 & 0.426 
& 0.407 & 0.385  & \textcolor{red}{0.402} & \textcolor{blue}{0.384}
& 0.465 & 0.517
& 0.477 & 0.539 \\
Covid
& \textcolor{red}{0.049} & \textcolor{blue}{1.383} 
& 0.066 & 2.135
& 0.053 & 1.539 & \textcolor{red}{0.051} & \textcolor{blue}{1.436}
& 0.059 & 1.722
& 0.063 & 1.995\\
ECL
& 0.268 & 0.168 
& 0.277 & 0.184
& \textcolor{red}{0.255} & \textcolor{blue}{0.158} & \textcolor{red}{0.265} & \textcolor{blue}{0.169}
& 0.274 & 0.177
& 0.269 & 0.181 \\
NN
& \textcolor{red}{0.617} & \textcolor{blue}{0.804} 
& 0.630 & 0.821 
& 0.625 & 0.828 & \textcolor{red}{0.659} & \textcolor{blue}{0.865}
& 0.672 & 0.913
& 0.661 & 0.884\\
\midrule
\rowcolor{gray!35}\multicolumn{13}{c}{\textbf{Out-of-Domain Test}} \\
\midrule
Wind
& \textcolor{red}{0.737} & \textcolor{blue}{1.015}& 0.779 & 1.655
& 0.769 & 1.447 & \textcolor{red}{0.722} & \textcolor{blue}{0.963}
& 0.755 & 1.324
& 0.772 & 1.564 \\
Solar
& \textcolor{red}{0.295} & \textcolor{blue}{0.228}& 0.361 & 0.293
& 0.349 & 0.287 & \textcolor{red}{0.322} & \textcolor{blue}{0.274}
& 0.341 & 0.297
& 0.353 & 0.311 \\
AQShunyi
& \textcolor{red}{0.445} & \textcolor{blue}{0.612}& 0.464 & 0.679
& 0.451 & 0.638 & \textcolor{red}{0.472} & \textcolor{blue}{0.688}
& 0.487 & 0.706
& 0.493 & 0.717 \\
CzenLan
& \textcolor{red}{0.308} & \textcolor{blue}{0.274}& 0.367 & 0.342
& 0.359 & 0.311 & \textcolor{red}{0.319} & \textcolor{blue}{0.288}
& 0.336 & 0.301
& 0.355 & 0.337\\
ZafNoo
& 0.458 & 0.547& 0.447 & 0.548
& \textcolor{red}{0.430} & \textcolor{blue}{0.512} & 0.471 & 0.583
& \textcolor{red}{0.465} & \textcolor{blue}{0.574}
& 0.490 & 0.622 \\
NASDAQ
& \textcolor{red}{0.635} & \textcolor{blue}{0.735}& 0.936 & 1.955
& 0.922 & 1.854 & \textcolor{red}{0.640} & \textcolor{blue}{0.749}
& 0.672 & 0.793
& 0.715 & 1.023 \\
PEMS
& \textcolor{red}{0.301} & \textcolor{blue}{0.256}& 0.321 & 0.270
& 0.317 & 0.275& 0.332 & 0.291
& \textcolor{red}{0.314} & \textcolor{blue}{0.266}
& 0.328 & 0.281  \\
\bottomrule
\end{tabular}}}
\label{tab:all}
\end{table*}

\subsection{Ablation Setup}
To investigate the impact of LLM parameters in TSF,
we consider three model configurations(original architecture and two ablation variants, as shown in Fig.~\ref{fig:wo}) with different levels of reliance on pretrained LLMs:
(1) \emph{\textbf{w/ pre-training}}, where the LLM serves as the backbone model and retains its pretrained weights;
in this setting, the LLM components are frozen in {LLM4TSF(Pre-align)} and fine-tuned in {LLM4TSF(Post-align)};
(2) \emph{\textbf{w/o pre-training}}, which follows the same architecture as the pretrained setting
but randomly initializes all LLM parameters;
and (3) \emph{\textbf{w/o LLM}}, where the LLM components are entirely removed from the architecture, retaining only other non-LLM modules. 
The three architectures are trained following the same procedure as in Sec.~\ref{Experimental Setup} and evaluated on a lot of scenarios.

\subsection{Ablation Results}
\textbf{Main results}. As shown in Tab.~\ref{tab:all}, when comparing the {w/ pre-training} against the two ablated counterparts ({w/o pre-training} and {w/o LLM}), LLM4TSF(Pre-align) with pretrained LLM parameters achieves the lowest forecasting error on \emph{\textbf{6/10}} datasets in the in-domain test and \emph{\textbf{6/7}} datasets in the out-of-domain test. Similarly, LLM4TSF(Post-align) with pre-training attains the lowest error on 7/10 in-domain and 5/7 out-of-domain datasets. To better understand the role of LLMs in TSF, we disentangle the effects of LLM, to distinguish whether gains arise from pretrained semantic priors or merely from architectural modeling capacity.

\textbf{(I) w/ pre-training vs. w/o pre-training.}\quad
we observe that LLM4TSF(Pre-align) consistently benefits from LLM prior knowledge: training with pretrained LLM parameters outperforms random initialization on all datasets (i.e., \emph{\textbf{10/10}} in-domain and \emph{\textbf{7/7}} out-of-domain test). For LLM4TSF(Post-align), models {w/ pre-training} perform better on \emph{\textbf{7/10}} in-domain datasets and \emph{\textbf{5/7}} out-of-domain test. 
\textbf{(II) w/o pre-training vs. w/o LLM.}\quad
For LLM4TSF(Pre-align), retaining a randomly initialized LLM consistently underperforms directly removing the LLM backbone across all tests (\emph{\textbf{10/10}} in-domain and \emph{\textbf{7/7}} out-of-domain). In contrast, for LLM4TSF(Post-align), retaining a randomly initialized LLM outperforms removing the LLM backbone on \emph{\textbf{7/10}} in-domain tests and on \emph{\textbf{all}} out-of-domain tests. 

\textbf{Takeaways.}\quad The gains of LLM4TSF models arise from both LLM parameters and architectural capacity.
Moreover, we identify two interesting findings: \emph{\textbf{(1) across different alignment strategies, pre-trained priors contribute to performance to varying extents;}} and \emph{\textbf{(2) freezing LLM causes randomly initialized models to collapse, therefore under post-alignment, fully trainable LLMs can be optimized from scratch and outperform w/o LLM.}}

%% file: Section/05_need.tex
\section{Understanding When LLM4TSF Works}
In Sec.~\ref{Do the Gains Really Come from LLM Parameters?}, We observe that in 71\% of the in-domain and out-of-domain tests, the {w/ pre-training} setting consistently outperforms both {w/o pre-training} and {w/o LLM}. Nevertheless, a small number of datasets exhibit failure cases where such benefits do not materialize. Accordingly, in this section, we first analyze \emph{\textbf{(i) why LLMs may fail under certain TSF scenarios}}, and investigate \emph{\textbf{(ii) how to better leverage LLM strengths to maximize their effectiveness in TSF}}.

\subsection{Exploring LLM Preferences over TS Distributions}
\textbf{Differences in the statistical properties of TS.}\quad TS data exhibit substantial differences in statistical properties, leading to diverse modeling~\cite{wangtowards,siru2025time,cini2025relational}. On the one hand, TS often contain prominent global structural characteristics, such as long-term trends and seasonal patterns, which describe stable regularities over extended time horizons~\cite{wang2025timecf,majeedi2025lets}. On the other hand, TS exhibit local dynamic variations, including distribution shifting, changes in stationarity, and regime transitions~\cite{masserano2024enhancing,hu2025timefilter}, which reflect the evolving of temporal data. 

\textbf{Dataset-level statistical analysis.}\quad We analyze the statistical properties of all datasets used for evaluation from five perspectives, as shown in Fig.~\ref{fig:dataset_attribute_bars}. 
\textbf{(I) in-domain datasets.} In terms of shifting, ETTh2, ETTm2, Weather, Exchange, Covid, and NN exhibit  high shifting values, indicating pronounced changes in data distributions over time. The transition shows that ETTh2, ETTm2, Weather, Exchange, and Covid experience more frequent and complex state changes, reflecting multi-stage and multi-pattern dynamics. Regarding stationarity, Exchange and Covid display lower stationarity levels (larger values indicate poorer stationarity). However, seasonality and trend are balanced across datasets. \textbf{(II) out-of-domain datasets.} NASDAQ exhibits an exceptionally high shifting value. In terms of stationarity, both CzenLan and NASDAQ show low stationarity levels. By contrast, the remaining properties—transition, seasonality, and trend, are relatively balanced across datasets. More details are provided in Appendix~\ref{app_Statistical Properties of Dataset}.

\begin{figure}[t]
    \centering
    \vspace{-0.2cm}
    \includegraphics[width=\linewidth]{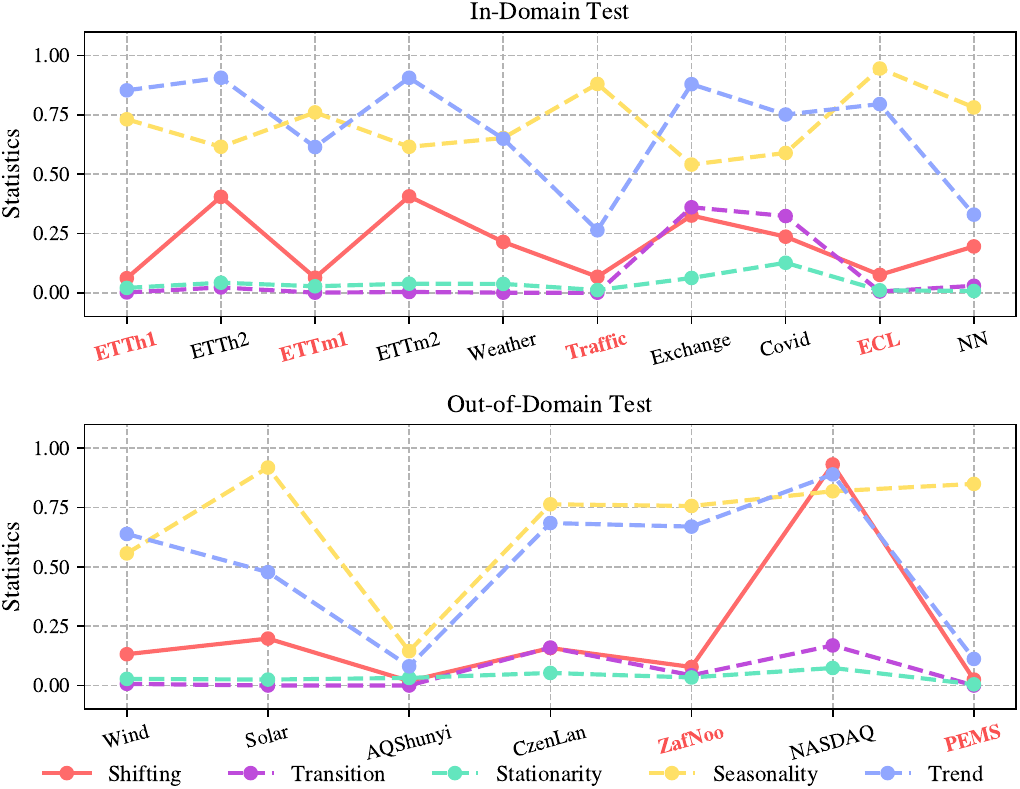}
    \caption{Statistical properties of different datasets. Datasets highlighted in \textcolor{red}{red} indicate cases where the {w/ pre-training} setting underperforms the corresponding ablation baselines in test.}
    \label{fig:p_mae}
    \vspace{-0.6cm}
\end{figure}

\begin{figure}[t]
    \centering
    \vspace{-0.0cm}
    \includegraphics[width=\linewidth]{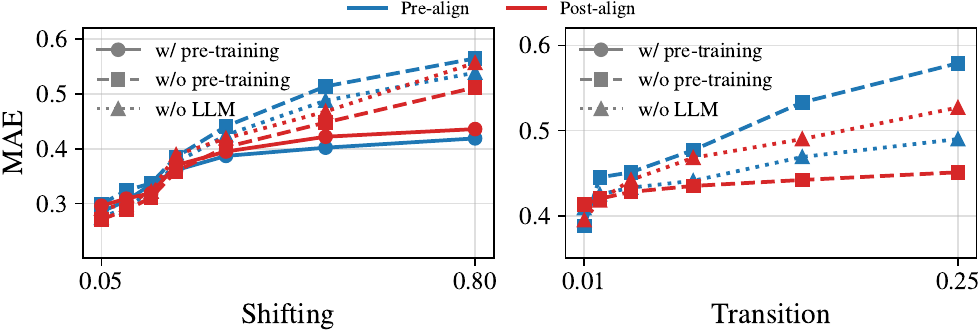}
    \caption{As shifting and transition increase, TSF becomes more challenging with higher MAE, while the advantage of LLMs becomes evident. Moreover, even w/o pre-training, a trainable LLM backbone outperforms non-LLM under high-transition regimes.}
    \label{fig:effect_shifting_transition}
    \vspace{-0.2cm}
\end{figure}

\textbf{Interactions between properties and performance.}\quad 
\emph{\textbf{Based on the statistical analysis, we observe a clear association between the TS properties and the model performance.}} When a TS exhibits strong shifting, pretrained LLM parameters tend to yield more pronounced gains. In contrast, when shifting is weak, satisfactory performance can often be achieved using only the encoder and decoder modules, as shown in Fig.~\ref{fig:p_mae}.
Moreover, under the post-alignment setting, where LLM parameters are updated during training, for datasets with high transition, the LLM from scratch can still outperform completely removing the LLM backbone. 

Since the statistical properties of real-world datasets are often entangled, we adopt a synthetic data generation approach with controlled decoupling to investigate the individual effects of shifting and transition on forecasting error, as shown in Fig.~\ref{fig:effect_shifting_transition}. More details are provided in Appendix~\ref{app_Impact of Statistical Properties} \& Appendix~\ref{app_Synthetic TS Generation}. As shifting increases, LLMs exhibit advantages; likewise, higher transition amplifies the performance gap between the w/o pre-training and w/o LLM.

\begin{figure*}[t]
    \centering
    \includegraphics[width=\textwidth]{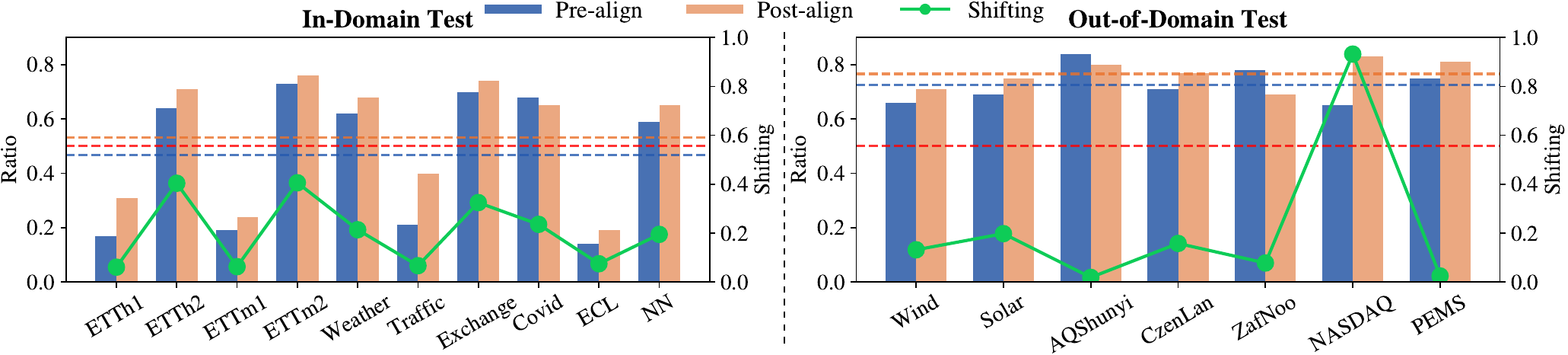}
    \caption{Relationship between the token ratio passing through the LLM (averaged over horizons $\{96, 192, 336, 720\}$) and shifting across different datasets. 
The \textcolor{blue}{"- - -"} \& \textcolor{orange}{"- - -"} indicate the average ratios of each dataset, and the \textcolor{red}{"- - -"} denotes the reference value of \textcolor{red}{0.5}.}
\label{fig:ratio_shifting}
\end{figure*}

\textbf{Takeaways.}\quad We find that shifting and transition play distinct roles in determining the effectiveness of LLM backbone in TSF:
\emph{\textbf{(1) when shifting is strong, pretrained LLM parameters are more likely to provide meaningful performance benefits;}}
\emph{\textbf{(2) when transition is high, a trainable Transformer backbone, even without pretrained LLM parameters, outperforms the w/o LLM variant.}}

\subsection{Routing Analysis: Pass Through the LLM or Skip?}
Previous analyses show the benefits of LLMs in TSF are scenario-dependent.
We therefore examine, when LLMs are more likely to play an active role.
Specifically, we employ a routing mechanism to assign path preferences to individual TS segments, referring to Appendix~\ref{app_Router Mechanism}.
As these routing decisions involve non-differentiable operations such as argmax or one-hot sampling, we use the Gumbel-Softmax reparameterization technique to facilitate optimization~\cite{gumbel1954statistical,jang2017categoricalreparameterizationgumbelsoftmax}, as shown in Fig.~\ref{fig:router}. Based on this, we pose the following four research questions (RQs):

\begin{figure}[t]
\vspace{-0.3cm}
    \centering
    \includegraphics[width=\linewidth]{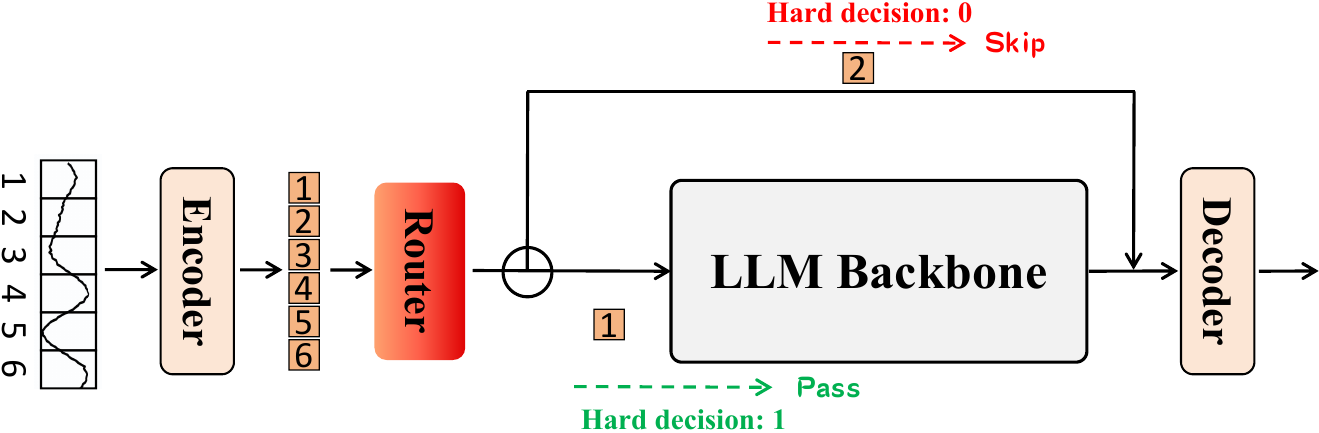}
    \caption{Routing analysis illustration. For each TS token, a trained router decides whether to route it to the LLM or skip. The LLM backbone is either w/ pre-training or w/o pre-training.}
    \label{fig:router}
\end{figure}

\begin{figure}[t]
    \centering
    \vspace{-0.3cm}
    \includegraphics[width=\linewidth]{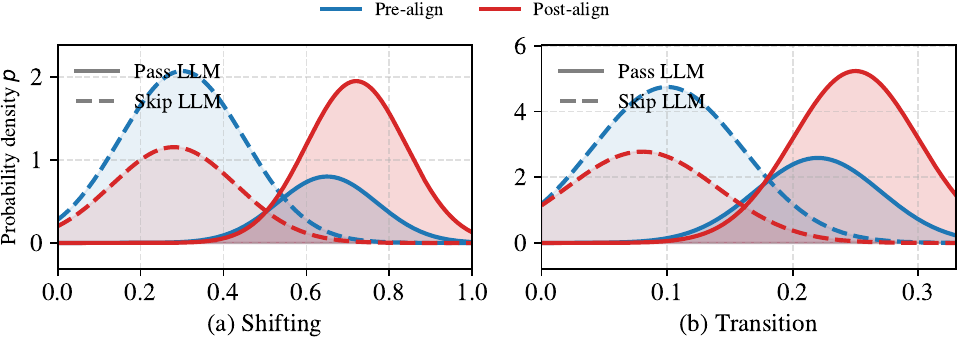}
    \caption{(a) With a pre-trained LLM backbone, tokens of passing LLM are concentrated in high shifting regions. 
(b) With pre-training, tokens of passing LLM are concentrated in high transition.}
    \label{fig:distribution_shifting_transition}
    \vspace{-0.5cm}
\end{figure}

\begin{table}[t]
\vspace{-0.1cm}
\centering
\caption{Impact of pretrained LLM parameters on the token passing ratio through the LLM, averaged over horizons $96, 192, 336, 720$.}
\renewcommand{\arraystretch}{1.00}
\resizebox{0.98\linewidth}{!}{
\begin{tabular}{c|cc|c|cc|c}
\toprule
\multirow{2}{*}{\textbf{}} 
& \multicolumn{3}{c|}{\textbf{LLM4TSF (Pre-align)}} 
& \multicolumn{3}{c}{\textbf{LLM4TSF (Post-align)}} \\
& \textbf{\makecell{w/ Pre-\\ training}} 
& \textbf{\makecell{w/o Pre-\\ training}} 
& \textbf{Change} 
& \textbf{\makecell{w/ Pre-\\ training}} 
& \textbf{\makecell{w/o Pre-\\ training}} 
& \textbf{Change} \\
\midrule

\rowcolor{gray!35}\multicolumn{7}{c}{\textbf{In-Domain Test}} \\
\midrule
ETTh1    & \textbf{17\%} & \textbf{24\%} & \textcolor{myred}{+7\%}  & \textbf{31\%}& 57\% & \textcolor{myred}{+26\%} \\
ETTh2    & 64\% & \textbf{43\%} & \textcolor{mygreen}{-21\%} & 71\% & 62\% & \textcolor{mygreen}{-9\%} \\
ETTm1    & \textbf{19\%} & \textbf{14\%} & \textcolor{mygreen}{-5\%}  & \textbf{24\%} & 66\% & \textcolor{myred}{+42\%} \\
ETTm2    & 73\% & \textbf{28\%} & \textcolor{mygreen}{-45\%} & 76\% & 68\% & \textcolor{mygreen}{-8\%} \\
Weather  & 62\% & \textbf{39\%} & \textcolor{mygreen}{-23\%} & 68\% & 62\% & \textcolor{mygreen}{-6\%} \\
Traffic  & \textbf{21\%} & \textbf{27\%} & \textcolor{myred}{+6\%}   & \textbf{40\%} & \textbf{46\%} & \textcolor{myred}{+6\%} \\
Exchange & 70\% & \textbf{31\%} & \textcolor{mygreen}{-39\%} & 74\% & 59\% & \textcolor{mygreen}{-15\%} \\
Covid    & 68\% & \textbf{15\%} & \textcolor{mygreen}{-53\%} & 65\% & 55\% & \textcolor{mygreen}{-10\%} \\
ECL      & \textbf{14\%} & \textbf{21\%} & \textcolor{myred}{+7\%}   & \textbf{19\%} & \textbf{37\%} & \textcolor{myred}{+18\%} \\
NN       & 59\% & \textbf{26\%} & \textcolor{mygreen}{-23\%} & 65\% & \textbf{43\%} & \textcolor{mygreen}{-22\%} \\

\midrule
\rowcolor{gray!35}\multicolumn{7}{c}{\textbf{Out-of-Domain Test}} \\
\midrule
Wind     & 66\% & \textbf{14\%} & \textcolor{mygreen}{-52\%} & 71\% & 65\% & \textcolor{mygreen}{-6\%} \\
Solar    & 69\% & \textbf{18\%} & \textcolor{mygreen}{-51\%} & 75\% & 60\% & \textcolor{mygreen}{-15\%} \\
AQShunyi & 84\% & \textbf{25\%} & \textcolor{mygreen}{-59\%} & 80\% & 58\% & \textcolor{mygreen}{-22\%} \\
CzenLan  & 71\% & \textbf{21\%} & \textcolor{mygreen}{-50\%} & 77\% & 71\% & \textcolor{mygreen}{-6\%} \\
ZafNoo   & 78\% & \textbf{26\%} & \textcolor{mygreen}{-52\%} & 69\% & 79\% & \textcolor{myred}{+10\%} \\
NASDAQ   & 65\% & \textbf{19\%} & \textcolor{mygreen}{-46\%} & 83\% & 75\% & \textcolor{mygreen}{-8\%} \\
PEMS     & 75\% & \textbf{15\%} & \textcolor{mygreen}{-60\%} & 81\% & 85\% & \textcolor{myred}{+4\%} \\
\bottomrule

\end{tabular}}
\begin{tablenotes}
\tiny
    \item *\textbf{Ratio \textless 50\%} indicates a stronger tendency to skip the LLM.
\end{tablenotes}
\label{tab:ratio_woLLM}
\vspace{-0.3cm}
\end{table}

\begin{figure}[t]
\vspace{-0.3cm}
    \centering
    \includegraphics[width=\linewidth]{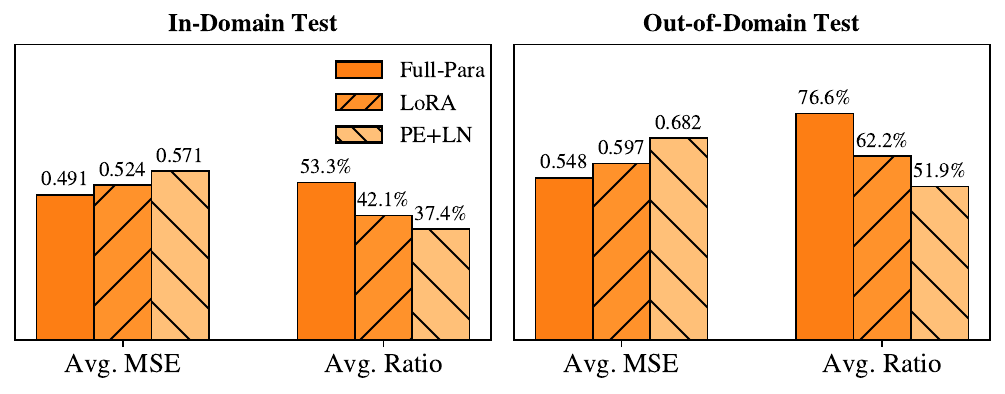}
    \caption{LLM4TSF(Post-align) performance and token passing ratios through the LLM under three training strategies: full-parameter fine-tuning, LoRA fine-tuning, and training PE+LN.}
    \label{fig:strategy}
    \vspace{-0.4cm}
\end{figure}

\emph{\textbf{RQ1: What properties of TS are likely to rely on LLMs?}}

We train LLM4TSF (Pre-align) \& (Post-align) under a cross-dataset learning setting, where each token is routed to either pass through or skip the LLM. As shown in Fig.~\ref{fig:ratio_shifting}, datasets with stronger shifting or unseen (out-of-domain) distributions tend to exhibit higher ratios of tokens passing through the LLM. Moreover, the Post-align setting, which fine-tunes LLM parameters, leads to higher passing ratios compared to Pre-align. Since different TS segments within the same dataset may exhibit distinct properties, dataset-level analysis alone is insufficient to fully characterize the router’s decision behavior. Therefore, we analyze the relationship between the properties of all test-set TS segments and their passing ratios to the LLM. Specifically, we visualize the joint density $p(x, D)$, where $x$ denotes properties (e.g., shifting or transition) and $D \in \{\text{pass}, \text{skip}\}$ indicates the routing decision. Each density curve is normalized such that its integral corresponds to the routing ratio $\mathbb{P}(D=d)$, as shown in Fig.~\ref{fig:distribution_shifting_transition}. More results are provided in Appendix~\ref{app_Impact of Statistical Properties}.

\emph{\textbf{RQ2: Do pre-trained parameters influence dependency?}}

We compare models with pretrained LLM parameters against counterparts with randomly initialized LLMs, assessing how pretraining alters the model’s tendency to use or skip LLM. As shown in Tab.~\ref{tab:ratio_woLLM}, datasets with higher shifting (e.g., ETTh2, ETTm2, Weather, and Exchange) tend to pass through the LLM, and the passing ratio decreases under the w/o pre-training setting; in contrast, datasets with lower shifting (e.g., ETTh1, ETTm1, Traffic, and ECL) exhibit the opposite trend. Moreover, under the LLM4TSF (Post-align) setting, datasets with lower transition(e.g., Traffic, ECL, and NN) show a strong tendency to skip the LLM, suggesting that for TS data with simpler transition patterns, the encoder \& decoder modules alone may suffice for forecasting. Overall, the results in Tab.~\ref{tab:all}, together with the routing statistics in Tab.~\ref{tab:ratio_woLLM}, indicate that tokens tend to select the path associated with lower forecasting error, with shifting and transition emerging as key factors underlying path selection behavior. 

\emph{\textbf{RQ3: Do training strategies modulate the LLMs?}}
By comparing different training strategies, we analyze whether they alter the overall tendency to use or skip LLM. Taking full-parameter fine-tuning as the baseline, we consider LoRA~\cite{hu2022lora} and a lightweight strategy that updates only positional embeddings and layer normalization, following prior work~\cite{zhou2023one,ICLR2025_e1de63ec}. Fig.~\ref{fig:strategy} show that Full-Para enables the most effective utilization of the LLM, yielding the best overall performance.
This indicates that unrestricted parameter optimization may better exploit the model’s representational capacity, thereby improving task performance.

\emph{\textbf{RQ4: Do stronger LLM consistently lead to greater gains?}}
We replace the LLM backbone  with more stronger Qwen-3~\cite{yang2025qwen3technicalreport} and observe that stronger general capabilities do not consistently translate into improved performance, as shown in Tab.~\ref{tab:size}. However, we find that artificially truncating the model by retaining only half of its layers, as in prior work~\cite{jin2023time,liu2024autotimes,liu2025timecma,ICLR2025_e1de63ec}, leads to noticeable performance degradation (see Appendix~\ref{app_Impact of Model Completeness}). More importantly, omitting prompts weakens the effectiveness of LLMs and leads to noticeable performance degradation, an effect that is particularly pronounced in out-of-domain scenarios.
It suggests that enriching prompt information is more impactful than indiscriminately scaling up the model backbone in TSF.

%% file: Section/06_dis.tex
\section{Discussion and Conclusion}
\textbf{Discussion.}\quad
\textbf{\emph{(1) Cross-dataset learning is a crucial prerequisite for unlocking the full potential of LLMs.}}
By exposing LLMs to diverse data, cross-dataset learning enables stronger performance.
\textbf{\emph{(2) Pre-alignment provides a more effective integration strategy for LLM4TSF.}}
Aligning TS inputs with word embeddings to lead to more compatible representations, resulting in lower errors.
\textbf{\emph{(3) The advantages of LLM4TSF arise from both pretrained knowledge and architectural modeling capacity.}}
They both contribute to improved forecasting capability.
\textbf{\emph{(4) LLM4TSF exhibit inherent preferences toward certain TS properties.}}
In particular, LLM tend to achieve advantages on TS with pronounced shifting or complex transition patterns.
\textbf{\emph{(5) Both a complete architecture and sufficient parameter optimization are essential for achieving strong performance.}}
Preserving the full LLM architecture and enabling adequate parameter optimization are necessary to leverage the LLMs.
\textbf{\emph{(6) The routing mechanism provides direct evidence for the observed macroscopic performance.}}
The routing offer an view of how LLM4TSF allocates modeling capacity, serving as a concrete explanation.
\textbf{\emph{(7) Blindly scaling up LLM backbones does not necessarily lead to better performance.}}
Simply increasing model size without adequate modality alignment may yield diminishing returns, limiting the benefits. More analysis are provided in Appendix~\ref{app_Analysis}.

\textbf{Conclusion.}\quad
In this work, we revisit the role of LLM4TSF and provide a clear characterization of when and why LLM4TSF models are effective. In addition, we show that the benefits of LLMs are neither universal nor incidental, but arise from the interplay between LLM knowledge and architectural capacity under specific distributions. By  fine-grained routing analysis, we reveal that LLMs exhibit consistent preferences over TS with different statistical properties, which explains their empirical gains and limitations. We not only clarify ongoing debates surrounding LLM4TSF, but also offer guidance for designing more principled models.

%% file: Section/related_work.tex
\section{Related Work}
\subsection{LLMs for TSF}
Directly applying fully pretrained LLMs to TSF tasks poses a central challenge in achieving effective modality alignment, as LLMs are originally optimized for discrete textual tokens rather than continuous temporal signals. Bridging the representational gap between TS data and the linguistic embedding space of LLMs has therefore become a key research focus~\cite{jiang2024empowering,wang2024news,liu2024autotimes,liu2025towards,jiang2025explainable}. Existing studies addressing this challenge can be broadly categorized into two representative paradigms, depending on whether alignment is performed before or after the LLM is involved in the modeling pipeline~\cite{zhou2025balm,xiong2025beyond,hu2025sst,tao2025values}.
The first category follows a pre-alignment strategy, which aims to align TS and textual modalities prior to LLM input. In this paradigm, raw or encoded TS are transformed into intermediate representations that are compatible with the LLM token embedding space, often via learnable projection layers or modality-specific encoders. These TS representations are then concatenated or interleaved with textual prompts and fed into a frozen LLM for downstream forecasting tasks. By keeping the LLM parameters fixed, this approach preserves the pretrained linguistic and semantic knowledge of the LLM while enabling it to process TS information in a unified embedding space. As a result, pre-alignment methods offer strong parameter efficiency and stability, making them particularly attractive in low-resource or deployment-constrained scenarios~\cite{zhang2024large,ceperic2024transforming}.
The second category adopts a post-alignment strategy, in which LLMs are fine-tuned to adapt to TSF tasks by reducing the representational discrepancy between textual and TS modalities within the LLM embedding space~\cite{bian2024multi,lee2025timecap,sun2025enhancing}. Rather than enforcing compatibility before input, these methods rely on task-driven learning signals to implicitly or explicitly align modalities during training. This is commonly achieved through joint optimization objectives, cross-modal alignment losses, or auxiliary supervision that encourages coherent representations across modalities. While the original LLM architecture is typically preserved, a subset of model parameters is updated to improve task-specific performance. Consequently, post-alignment approaches offer greater flexibility and expressive power at the cost of increased training complexity and computational overhead.

\subsection{Transfer Learning}
Transfer learning has become a widely adopted paradigm in deep learning, demonstrating remarkable effectiveness across a wide range of domains~\cite{weiss2016survey,jain2023data,wang2024f,zhang2024exploring,liu2025multitask}. By pretraining models on large-scale datasets and transferring them to downstream tasks, prior work has shown substantial improvements in both performance and data efficiency. In vision–language research, for example, pretrained multimodal models jointly learn representations from images and text, enabling effective transfer to applications such as image retrieval~\cite{wang2025generative,huynh2025collm}, image captioning~\cite{yu2022coca,huang2024solution}, and visual question answering~\cite{lin2024video,huynh2025visual}. In speech and language modeling, pretraining strategies that integrate acoustic signals with textual supervision have significantly advanced automatic speech recognition and semantic understanding. Similar successes have also been observed in domains such as healthcare~\cite{moon2022multi,yu2025umitunifyingmedicalimaging} and remote sensing~\cite{kuckreja2024geochat,zhan2025skyeyegpt}.
These advances suggest that when different modalities exhibit strong semantic complementarity, large-scale pretrained models—particularly LLMs, can learn representations with high transferability and generalization capability. However, this assumption does not directly extend to TS data and TSF tasks. Unlike images or speech signals, TS data represent real-world processes in a highly abstract numerical form, often lacking explicit semantic grounding in natural language. As a result, the correspondence between TS and textual modalities is inherently ambiguous, making it difficult to directly transfer pretrained knowledge from LLMs to TSF.
At the same time, generalization—especially under cross-domain settings—is a fundamental requirement of practical TSF systems. In many applications, the target domain may differ  from the source domain in terms of data distributions, temporal patterns, or underlying dynamics, while labeled TS data in the target domain are often limited or unavailable. Models trained with strong domain-specific inductive biases therefore tend to suffer significant performance degradation when deployed across domains.
In this context, LLMs offer a promising opportunity for cross-domain TSF due to their strong generalization abilities.  

\clearpage

%% file: Section/APP_analy.tex
\section{Analysis}
\label{app_Analysis}
\textbf{\emph{(1) Cross-dataset learning is a crucial prerequisite for unlocking the full potential of LLMs.}}\quad
Compared to small-scale training on a single dataset, diverse cross-dataset learning within a unified framework more effectively exploits the capabilities of LLMs. This strategy not only alleviates overfitting but also leads to stronger overall performance. Specifically, it surpasses single-dataset baselines on in-domain tasks and consistently outperforms a range of strong time-series foundation models as well as LLM-based approaches in out-of-domain evaluations, with the performance gains becoming increasingly pronounced as the diversity and scale of training data grow, as shown in Fig.~\ref{fig:data_ratio_mae_mse}.

\textbf{\emph{(2) Pre-alignment provides a more effective integration strategy for LLM4TSF.}}\quad
We observe that aligning TS inputs with word embeddings {before} feeding them into LLMs yields lower forecasting errors than performing alignment between text and TS representations within the LLM space. This finding suggests that pre-alignment enables a more compatible input representation, leading to more effective utilization of the pretrained LLM parameters.

\textbf{\emph{(3) The advantages of LLM4TSF arise from both pretrained knowledge and architectural modeling capacity.}}\quad
Our analysis shows that the performance improvements of LLM4TSF models stem from two complementary sources. Pretrained parameters endow the model with rich prior knowledge, while the expressive Transformer-based architecture provides strong sequence modeling capacity. Together, these factors enable LLM4TSF to achieve strong performance across a wide range of forecasting scenarios.

\textbf{\emph{(4) LLM4TSF exhibit inherent preferences toward certain TS properties.}}\quad
On the one hand, pretrained knowledge provides strong priors that are particularly beneficial when data distributions shift substantially over time or when models are evaluated on previously unseen out-of-domain datasets. On the other hand, the expressive architecture of LLMs offers strong modeling capacity, enabling them to better capture complex temporal dynamics characterized by frequent or abrupt transitions in underlying patterns. In contrast, factors such as stationarity, seasonality, or trends are not the primary driver, suggesting that simpler models may already be sufficient when such characteristics dominate.

\textbf{\emph{(5) Both a complete architecture and sufficient parameter optimization are essential for achieving strong performance.}}\quad
Our results show that full-parameter fine-tuning consistently outperforms parameter-efficient alternatives such as LoRA or partial adaptation strategies (e.g., positional encoding and layer normalization tuning). Moreover, artificially truncating the LLM by retaining only its shallow layers leads to noticeable performance degradation, as it undermines the model’s ability to fully exploit its architectural depth and pretrained capacity, as shown in the Fig.~\ref{fig:truncation}. 

\textbf{\emph{(6) The routing mechanism provides direct evidence for the observed macroscopic performance.}}\quad
We not only observe performance gaps in overall MAE/MSE, but also gain fine-grained insight into model decision-making. Empirically, the model prefers paths that yield lower prediction errors.
This token-level routing preference closely aligns with macroscopic performance trends, offering direct  micro-level evidence for when and why incorporating LLMs leads to performance gains.

\textbf{\emph{(7) Blindly scaling up LLM backbones does not necessarily lead to better performance.}}\quad
Our experiments reveal that simply replacing the backbone with a larger LLM does not consistently yield performance improvements. One possible reason is the inherent modality gap between natural language and TS data, which makes direct alignment of increasingly large LLMs with TSF tasks more challenging. In addition, prior studies have shown that large-scale TS models often contain substantial internal redundancy~\cite{qiu2025governmanyunveilingfewlayerdominance,wilinski2025exploring}, which may limit the practical benefits of naive model scaling.

%% file: Section/APP_TSF.tex
\section{Time Series Forecasting Task}
\label{app_Time Series Forecasting Task}
\subsection{Problem Formulation}
The TSF task aims to predict future observations based on historical TS data.
Formally, given a multivariate TS of length $T$,
$$
\mathbf{X}_{1:T} = \{\mathbf{x}_1, \mathbf{x}_2, \ldots, \mathbf{x}_T\}, \quad \mathbf{x}_t \in \mathbb{R}^d,
$$
where $\mathbf{x}_t$ denotes the $d$-dimensional observation at time step $t$, the objective of TSF is to forecast the next $H$ time steps,
$$
\hat{\mathbf{X}}_{T+1:T+H} = \{\hat{\mathbf{x}}_{T+1}, \hat{\mathbf{x}}_{T+2}, \ldots, \hat{\mathbf{x}}_{T+H}\}.
$$

Each observation is first mapped into a latent representation space to facilitate downstream modeling. Specifically, the embedding at time step $t$ is defined as:
$
\mathbf{z}_t = f_{\mathrm{encoder}}(\mathbf{x}_t), \quad \mathbf{z}_t \in \mathbb{R}^k,
$
where $f_{\mathrm{encoder}}(\cdot)$ denotes the embedding function and $k$ is the latent dimensionality.
Based on the encoded sequence $\{\mathbf{z}_1, \mathbf{z}_2, \ldots, \mathbf{z}_T\}$, a forecasting model learns a mapping that captures temporal dependencies and generates predictions for future time steps.

\subsection{In-domain and Cross-domain Evaluation}
\label{app_In-domain and Cross-domain Evaluation}
We consider both in-domain and cross-domain evaluation settings for the TSF task, which differ in the relationship between the training and testing data distributions.

\paragraph{In-domain Evaluation.}
In the in-domain setting, the training and test TS are drawn from the same underlying data distribution.
Formally, let $\mathcal{D}_{\mathrm{train}}$ and $\mathcal{D}_{\mathrm{test}}$ denote the distributions of the training and test TS, respectively.
In-domain evaluation assumes
$
\mathcal{D}_{\mathrm{train}} = \mathcal{D}_{\mathrm{test}}.
$
Under this setting, the forecasting model is trained and evaluated on TS that share similar temporal patterns, statistical properties, and domain characteristics.
The goal is to assess the model’s ability to capture temporal dependencies within a fixed domain.

\paragraph{Cross-domain Evaluation.}
In contrast, the cross-domain setting evaluates the model’s generalization ability when the training and test TS originate from different domains.
Specifically, the training and test distributions satisfy
$
\mathcal{D}_{\mathrm{train}} \neq \mathcal{D}_{\mathrm{test}}.
$
These domains may differ in data distributions, temporal dynamics, scales, or underlying generative processes.
During training, the model has access only to TS sampled from $\mathcal{D}_{\mathrm{train}}$, while at test time it is required to perform TSF on unseen TS drawn from $\mathcal{D}_{\mathrm{test}}$.

\subsection{Zero-shot and Few-shot Test}
In cross-domain TSF, we consider zero-shot and few-shot test settings, which differ in the amount of target-domain supervision available at test time.
Let $\mathcal{D}_{\mathrm{src}}$ denote the source-domain TS distribution used for training, and $\mathcal{D}_{\mathrm{tgt}}$ denote the target-domain TS distribution used for testing, where
$
\mathcal{D}_{\mathrm{src}} \neq \mathcal{D}_{\mathrm{tgt}}.
$

\paragraph{Zero-shot Test.}
In the zero-shot setting, the forecasting model is trained solely on TS sampled from the source domain and is directly evaluated on the target domain without observing any labeled TS from $\mathcal{D}_{\mathrm{tgt}}$.
Formally, the model learns a forecasting function $f_{\theta}$ using training samples:
$
\{\mathbf{X}^{(i)}_{1:T}, \mathbf{X}^{(i)}_{T+1:T+H}\}_{i=1}^{N}
\sim \mathcal{D}_{\mathrm{src}},
$
and is evaluated on TS
$
\mathbf{X}_{1:T} \sim \mathcal{D}_{\mathrm{tgt}},
$
where the predicted future values are given by
$
\hat{\mathbf{X}}_{T+1:T+H} = f_{\theta}(\mathbf{X}_{1:T}).
$
This setting evaluates the model’s ability to generalize across domains purely through transferable representations learned during pretraining.

\paragraph{Few-shot Test.}
In the few-shot setting, the model is provided with a small labeled support set from the target domain.
Specifically, a support set
$
\mathcal{S}_{\mathrm{tgt}} =
\left\{
\left(\mathbf{X}^{(j)}_{1:T}, \mathbf{X}^{(j)}_{T+1:T+H}\right)
\right\}_{j=1}^{K},
 K \ll N,
$
is sampled from $\mathcal{D}_{\mathrm{tgt}}$.
The model adapts to the target domain by conditioning on $\mathcal{S}_{\mathrm{tgt}}$, yielding an adapted predictor
$
f_{\theta'} = \mathcal{A}(f_{\theta}, \mathcal{S}_{\mathrm{tgt}}),
$
where $\mathcal{A}(\cdot)$ denotes a lightweight adaptation mechanism, such as prompt-based conditioning or parameter-efficient tuning.
The adapted model is then evaluated on unseen TS from $\mathcal{D}_{\mathrm{tgt}}$.
Compared to zero-shot test, the few-shot setting assesses whether limited target-domain supervision can further improve forecasting performance, while still maintaining strong generalization across domains.

\clearpage

%% file: Section/APP_prompt.tex
\section{Text Prompt}
\label{app_Text Prompt}
\textbf{ETTh1 prompt.}\quad
The ETTh1 designed for time-series forecasting at 1-hour intervals, contains data points with the target variable "oil temperature" and six power load features. Given the past 512  observations, predict the next 96 time steps. The input window includes a minimum value of \{min value\}, a maximum value of \{max value\}, and a median value of \{median value\}.

\textbf{ETTh2 prompt.}\quad
The ETTh2 designed for time-series forecasting at 1-hour intervals, contains data points with the target variable "oil temperature" and six power load features. Given the past 512  observations, predict the next 96 time steps. The input window includes a minimum value of \{min value\}, a maximum value of \{max value\}, and a median value of \{median value\}.

\textbf{ETTm1 prompt.}\quad
The ETTm1 designed for time-series forecasting at 15-minutes intervals, contains data points with the target variable "oil temperature" and six power load features. Given the past 512  observations, predict the next 96 time steps. The input window includes a minimum value of \{min value\}, a maximum value of \{max value\}, and a median value of \{median value\}.

\textbf{ETTm2 prompt.}\quad
The ETTm2 designed for time-series forecasting at 15-minutes intervals, contains data points with the target variable "oil temperature" and six power load features. Given the past 512  observations, predict the next 96 time steps. The input window includes a minimum value of \{min value\}, a maximum value of \{max value\}, and a median value of \{median value\}.

\textbf{Weather prompt.}\quad
The Weather dataset is designed for time-series forecasting with data recorded every 10 minutes and contains 21 meteorological indicators, such as air temperature, humidity, and wind-related variables. Given the past {input length} observations, predict the next {prediction length} time steps. The input window includes a minimum value of \{min value\}, a maximum value of \{max value\}, and a median value of \{median value\}.

\textbf{Traffic prompt.}\quad
The Traffic dataset describes road occupancy conditions and contains hourly measurements collected from highway sensors. Given the past {input length} observations, predict the next {prediction length} time steps. The input window includes a minimum value of \{min value\}, a maximum value of \{max value\}, and a median value of \{median value\}.

\textbf{Exchange prompt.}\quad
The Exchange dataset contains daily exchange rate data for eight countries and is commonly used for time-series analysis and forecasting of currency fluctuations. Given the past {input length} observations, predict the next {prediction length} time steps. The input window includes a minimum value of \{min value\}, a maximum value of \{max value\}, and a median value of \{median value\}.

\textbf{ECL prompt.}\quad
The ECL dataset represents the hourly electricity consumption, recorded in kilowatts. Given the past {input length} observations, predict the next {prediction length} time steps. The input window includes a minimum value of \{min value\}, a maximum value of \{max value\}, and a median value of \{median value\}.

\textbf{NN prompt.}\quad
The NN dataset consists of 111 daily time series drawn from a homogeneous population of empirical cash demand data. Given the past {input length} observations, predict the next {prediction length} time steps. The input window includes a minimum value of \{min value\}, a maximum value of \{max value\}, and a median value of \{median value\}.

\textbf{Wind prompt.}\quad
The Wind dataset consists of time-series data recorded at 15-minute intervals, and includes variables such as wind speed, wind direction, temperature, pressure, humidity, and wind power. Given the past {input length} observations, predict the next {prediction length} time steps. The input window includes a minimum value of \{min value\}, a maximum value of \{max value\}, and a median value of \{median value\}.

\textbf{Solar prompt.}\quad
The Solar dataset contains 137 time series representing solar power production, recorded at 10-minute intervals. Given the past {input length} observations, predict the next {prediction length} time steps. The input window includes a minimum value of \{min value\}, a maximum value of \{max value\}, and a median value of \{median value\}.

\textbf{AQShunyi prompt.}\quad
The AQShunyi dataset includes 11 hourly time series capturing air quality and meteorological conditions in the Shunyi District of Beijing, with variables such as PM2.5, PM10, SO$_2$, NO$_2$, CO, O$_3$, temperature, air pressure, humidity, wind speed, and precipitation. Given the past {input length} observations, predict the next {prediction length} time steps. The input window includes a minimum value of \{min value\}, a maximum value of \{max value\}, and a median value of \{median value\}.

\textbf{Czelan prompt.}\quad
The Czelan dataset contains eight time series of plant sap flow measurements recorded at half-hour intervals. Given the past {input length} observations, predict the next {prediction length} time steps. The input window includes a minimum value of \{min value\}, a maximum value of \{max value\}, and a median value of \{median value\}.

\textbf{ZafNoo prompt.}\quad
The ZafNoo dataset consists of 11 time series of plant sap flow measurements, recorded at half-hour intervals. Given the past {input length} observations, predict the next {prediction length} time steps. The input window includes a minimum value of \{min value\}, a maximum value of \{max value\}, and a median value of \{median value\}.

\textbf{PEMS prompt.}\quad
The PEMS dataset includes 48 months of hourly data describing road occupancy rates measured by various traffic sensors. Given the past {input length} observations, predict the next {prediction length} time steps. The input window includes a minimum value of \{min value\}, a maximum value of \{max value\}, and a median value of \{median value\}.

\textbf{NASDAQ prompt.}\quad
The NASDAQ dataset encompasses companies listed on the NASDAQ stock exchange, with regularly updated time-series data excluding test listings. Given the past {input length} observations, predict the next {prediction length} time steps. The input window includes a minimum value of \{min value\}, a maximum value of \{max value\}, and a median value of \{median value\}.

%% file: Section/APP_baseline.tex
\section{Baselines}
\label{app_Baselines}
{Chronos}~\cite{ansari2024chronos} reformulates time series forecasting for Transformer-based architectures via a two-stage pre-processing strategy consisting of normalization and discretization. Specifically, each observation is normalized by the mean absolute value of its historical window to ensure scale consistency across different series. The normalized values are then quantized into a finite set of bins, converting the continuous time series into discrete token sequences. These tokens are subsequently modeled using a T5-style Transformer trained with a cross-entropy objective, allowing the model to learn generalizable TS representations while fully exploiting the strengths of sequence modeling frameworks.

{UniTS}~\cite{gao2024units} is a unified time series foundation model that supports a universal task formulation, enabling a wide range of TS tasks including forecasting, classification, imputation, and anomaly detection within a single framework. This capability is realized through a unified network backbone that integrates sequence-wise and variable-wise attention mechanisms together with a dynamic linear operator, allowing the model to flexibly capture temporal dependencies and inter-variable relationships. The entire architecture is trained end-to-end as a single unified model across tasks.
Extensive experiments conducted on 38 datasets spanning multiple domains demonstrate that {UniTS} consistently outperforms specialized task-specific models as well as repurposed natural language-based LLMs. 

{Moirai}~\cite{moirai} is a large-scale foundation model for TSF that departs from the conventional practice of training separate models for individual datasets. Through architectural enhancements to the standard Transformer, {Moirai} enables effective cross-frequency learning, accommodates arbitrary multivariate input dimensionalities, and adapts to heterogeneous data distributions across diverse domains. The model is trained on the Large-scale Open Time Series Archive, comprising more than 27 billion observations spanning nine domains. Owing to its large-scale pretraining, {Moirai} exhibits strong zero-shot forecasting performance, often matching or outperforming models that are fully fine-tuned on target datasets.

{UniTime}~\cite{liu2024unitime} is a unified foundation model for time series analysis that supports multiple tasks under a single modeling framework, including forecasting, classification, imputation, and anomaly detection. It introduces a unified task formulation together with a shared backbone architecture, allowing different TS tasks to be jointly learned without task-specific redesign. By leveraging a task-agnostic representation and a unified training objective, {UniTime} achieves strong generalization across tasks and datasets.
Extensive evaluations demonstrate that {UniTime} consistently outperforms task-specific baselines and exhibits robust zero-shot and few-shot performance when transferred to unseen datasets and tasks, highlighting its effectiveness as a general-purpose time series model.

{Time-LLM}~\citep{jin2023time} performs time series forecasting by reusing a pretrained large language model whose parameters remain entirely frozen. To adapt the model to TS inputs and outputs, it introduces two lightweight trainable modules: {Patch Reprogramming}, which converts TS segments into LLM-compatible representations, and {Output Projection}, which maps the model outputs to the forecasting space. In addition, the method follows a channel-independent formulation, decomposing multivariate forecasting problems into multiple parallel univariate prediction tasks, thereby enabling efficient adaptation without modifying the backbone LLM.
\clearpage

%% file: Section/APP_Dataset.tex
\section{Datasets}
\label{app_Datasets}
\subsection{General Overview}
\label{app_General Overview}
\begin{figure}[h]
    \centering
    \includegraphics[width=0.8\linewidth]{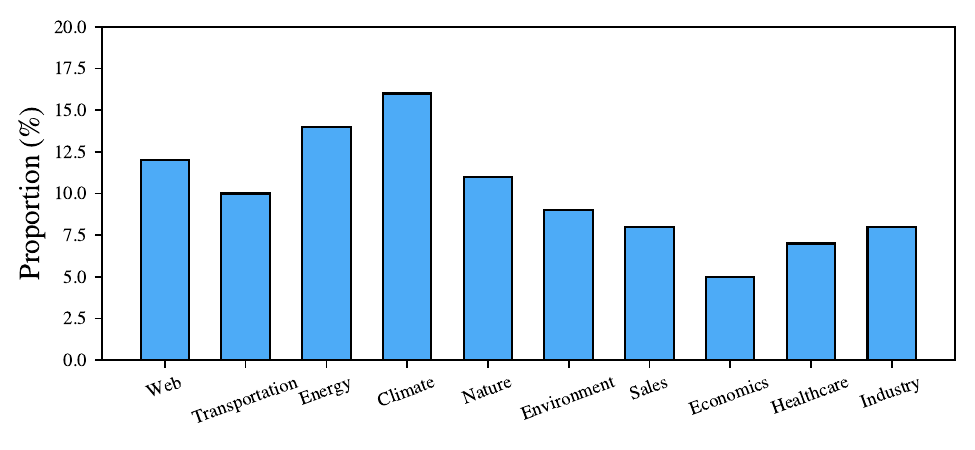}
    \caption{Proportional distribution of training datasets across ten real-world domains.}
    \label{fig:dataset_ratio}
\end{figure}

The training datasets cover a wide range of real-world domains, as shown in Fig.~\ref{fig:dataset_ratio}, including web, transportation, energy, nature, environment, climate, sales, economics, healthcare, and industry.
Such broad domain coverage introduces substantial diversity in temporal patterns, scales, seasonalities, and noise characteristics, which is crucial for learning transferable representations and improving generalization in cross-domain TSF settings.
In addition, the datasets used for in-domain and out-of-domain tests are listed separately, as summarized in Tab.~\ref{tab:datasets}.
Among them, 10 datasets, including ETT (four subsets), Weather, Traffic, Exchange, Covid, ECL, and NN,
are further split into training, validation, and test subsets.
The held-out test splits of these datasets are used for in-domain evaluation. For the ETT datasets, we follow the standard train--validation--test split ratio of 6:2:2,
while all other datasets in set~A are split using a 7:1:2 ratio.
Set~B contains 7 datasets, namely Wind, Solar, AQShunyi, CzenLan, ZafNoo, NASDAQ, and PEMS.
Set~B are completely excluded from the training process and are used solely for out-of-domain test.
Some of the collected datasets contain N/A or invalid values. To ensure data quality and training stability, we apply linear interpolation to impute missing values in all datasets.

\begin{table*}[h]
\centering
\caption{Summary of datasets used in experiments. The table reports the number of variables and timestamps for each time series dataset.}
\label{tab:datasets}
\begin{tabular}{ccc || ccc}
\toprule
\textbf{Dataset} & \textbf{Variables} & \textbf{Timestamps} &
\textbf{Dataset} & \textbf{Variables} & \textbf{Timestamps} \\
\midrule
ETTh1     & 7   & 14,400  & Wind     & 7   & 48,673  \\
ETTh2     & 7   & 14,400  & Solar    & 137 & 52,560  \\
ETTm1     & 7   & 57,600  & AQShunyi & 11  & 35,064  \\
ETTm2     & 7   & 57,600  & CzenLan  & 11  & 19,934  \\
Weather   & 21  & 52,696  & ZafNoo   & 11  & 19,225  \\
Traffic   & 862 & 17,544  & NASDAQ   & 5   & 1,244   \\
Exchange  & 8   & 7,588   & PEMS     & 170 & 17,856  \\
Covid     & 948 & 1,392   &          &     &        \\
ECL       & 321 & 26,304  &          &     &        \\
NN        & 111 & 791    &          &     &        \\
\bottomrule
\end{tabular}
\end{table*}

\subsection{Statistical Properties of Dataset}
\label{app_Statistical Properties of Dataset}
\begin{figure}[H]
    \centering
    \includegraphics[width=0.85\linewidth]{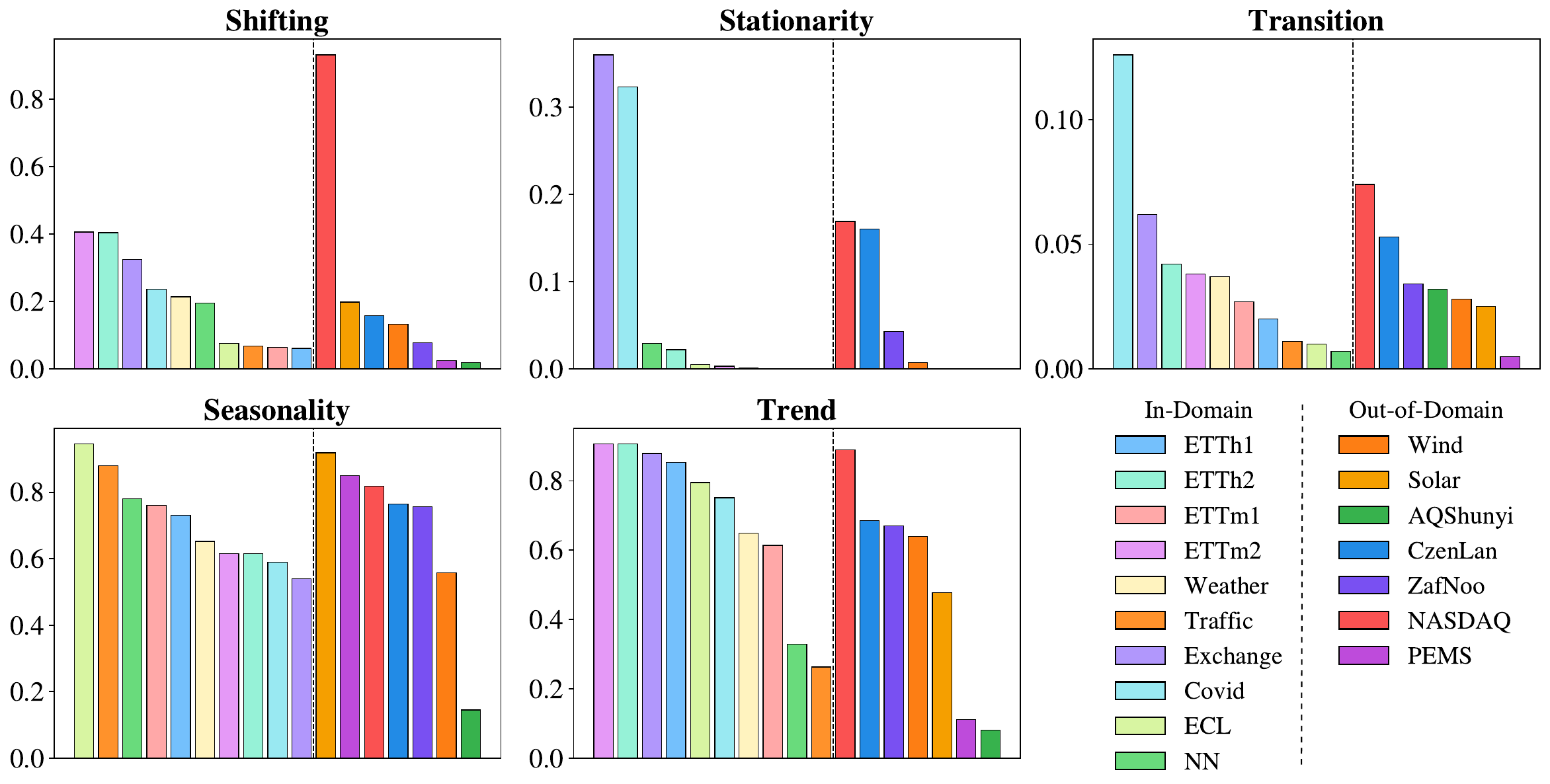}
    \caption{Analysis of the 17 in-domain and out-of-domain datasets from Shifting, Stationarity, Transition, Seasonality, and Trend.}
    \label{fig:dataset_attribute_bars}
\end{figure}

We visualize the statistical properties of all datasets used in testing, as shown in the Fig.~\ref{fig:dataset_attribute_bars}.
Shifting describes how the activation regions of the TS evolve along the temporal axis, capturing changes in distributional structure over time.
Stationarity reflects whether the statistical properties of a TS remain stable over time, indicating the presence or absence of distributional drift.
Transition characterizes the sequential dependency between symbolic states, reflecting the complexity of temporal transitions and local dynamical behavior.
Seasonality measures the strength of recurring temporal patterns at fixed intervals, which are commonly observed in real-world periodic processes.
Trend characterizes the long-term directional movement of the series, revealing persistent growth or decline patterns.
The corresponding computation procedures are summarized in Tab.~\ref{tab:shifting}--\ref{tab:trend}.

\begin{table}[H]
\centering
\caption{Algorithm: Shifting Computation}
\label{tab:shifting}
\resizebox{0.65\linewidth}{!}{
\begin{tabular}{ll}
\toprule
\multicolumn{2}{l}{\textbf{Input:} Time series $\mathbf{X} \in \mathbb{R}^{T \times 1}$} \\
\multicolumn{2}{l}{\textbf{Output:} Shifting $\delta \in (0,1)$} \\
\midrule
1: & Normalize $\mathbf{X}$ using z-score normalization to obtain $\mathbf{Z} \in \mathbb{R}^{T \times 1}$. \\

2: & Compute $Z_{\min} = \min(\mathbf{Z})$ and $Z_{\max} = \max(\mathbf{Z})$. \\

3: & Construct $m$ uniformly spaced value levels: \\
   & \hspace{1em}$
   \ell_i = Z_{\min} + \frac{i-1}{m-1}(Z_{\max}-Z_{\min}), \quad i=1,\ldots,m.
   $ \\

4: & \textbf{for} each level $\ell_i$ \textbf{do} \\

5: & \hspace{1em}Identify activated time indices
$
\mathcal{T}_i = \{\, t \mid Z_t > \ell_i,\; 1 \le t \le T \,\}.
$ \\

6: & \hspace{1em}Compute the temporal center
$
c_i = \mathrm{median}(\mathcal{T}_i).
$ \\

7: & Apply min--max normalization to $\{c_i\}_{i=1}^m$ to obtain $\{\tilde{c}_i\}_{i=1}^m$. \\

8: & \textbf{return}
$
\delta = \left| \mathrm{median}(\{\tilde{c}_1,\tilde{c}_2,\ldots,\tilde{c}_m\}) \right|.
$ \\
\bottomrule
\end{tabular}
}
\end{table}

\begin{table}[H]
\centering
\caption{Algorithm 2: Stationarity Computation}
\label{tab:stationarity}
\resizebox{0.7\linewidth}{!}{
\begin{tabular}{ll}
\toprule
\multicolumn{2}{l}{\textbf{Input:} Time series $\mathbf{X} = \langle x_1, x_2, \ldots, x_T \rangle \in \mathbb{R}^{T \times 1}$} \\
\multicolumn{2}{l}{\textbf{Output:} Stationarity indicator $\gamma \in \{0,1\}$ of $\mathbf{X}$} \\
\midrule
1: & Compute the Augmented Dickey--Fuller (ADF) statistic $s \leftarrow \mathrm{ADF}(\mathbf{X})$. \\

2: & \textbf{return}
$
\gamma =
\begin{cases}
1, & \text{if } s \le 0.05, \\
0, & \text{otherwise}.
\end{cases}
$ \\
\bottomrule
\end{tabular}
}
\end{table}

\begin{table}[H]
\centering
\caption{Algorithm: Transition Computation}
\label{tab:transition}
\resizebox{0.8\linewidth}{!}{
\begin{tabular}{ll}
\toprule
\multicolumn{2}{l}{\textbf{Input:} Time series $\mathbf{X} \in \mathbb{R}^{T \times 1}$} \\
\multicolumn{2}{l}{\textbf{Output:} Transition $\Delta \in \left(0, \tfrac{1}{3}\right)$} \\
\midrule
1: & Estimate the characteristic lag $\tau$ as the first zero-crossing point of the autocorrelation function of $\mathbf{X}$. \\

2: & Downsample $\mathbf{X}$ with stride $\tau$ to obtain a reduced sequence $\mathbf{Y} \in \mathbb{R}^{T' \times 1}$. \\

3: & Obtain the rank ordering of $\mathbf{Y}$ by computing the permutation index
$
\mathbf{r} = \mathrm{argsort}(\mathbf{Y}).
$ \\

4: & Discretize $\mathbf{Y}$ into a symbolic sequence $\mathbf{Z} \in \{0,1,2\}^{T'}$ via
$
Z_j = \left\lfloor \frac{3 \, r_j}{T'} \right\rfloor, \quad j = 1,\ldots,T'.
$ \\

5: & Initialize a transition count matrix $\mathbf{M} \in \mathbb{R}^{3 \times 3}$ with zeros. \\

6: & \textbf{for} $j = 1$ \textbf{to} $T'-1$ \textbf{do} \\
   & \hspace{1em}Increment the transition count
   $
   M_{Z_j,\, Z_{j+1}} \leftarrow M_{Z_j,\, Z_{j+1}} + 1.
   $ \\

7: & Normalize the transition matrix by sequence length:
$
\mathbf{M}' = \frac{1}{T'} \mathbf{M}.
$ \\

8: & Compute the covariance matrix $\mathbf{C}$ of the column vectors of $\mathbf{M}'$. \\

9: & \textbf{return}
$
\Delta = \mathrm{tr}(\mathbf{C}).
$ \\
\bottomrule
\end{tabular}
}
\end{table}

\begin{table}[H]
\centering
\caption{Algorithm: Seasonality Computation}
\label{tab:seasonality}
\resizebox{0.8\linewidth}{!}{
\begin{tabular}{ll}
\toprule
\multicolumn{2}{l}{\textbf{Input:} Time series $\mathbf{X} = \{x_1, x_2, \ldots, x_T\} \in \mathbb{R}^{T \times 1}$} \\
\multicolumn{2}{l}{\textbf{Output:} Seasonality value $\zeta \in (0,1)$} \\
\midrule
1: & Decompose $\mathbf{X}$ into seasonal, trend, and residual components using STL:
$
\mathbf{X} = \mathbf{S} + \mathbf{T} + \mathbf{R}.
$ \\

2: & Compute the variance of the residual component $\mathbf{R}$. \\

3: & Compute the variance of the combined seasonal and residual component $\mathbf{S} + \mathbf{R}$. \\

4: & \textbf{return}
$
\zeta = \max\!\left(0,\; 1 - \frac{\mathrm{var}(\mathbf{R})}{\mathrm{var}(\mathbf{S} + \mathbf{R})}\right).
$ \\
\bottomrule
\end{tabular}
}
\end{table}

\begin{table}[H]
\centering
\caption{Algorithm: Trend Computation}
\label{tab:trend}
\resizebox{0.8\linewidth}{!}{
\begin{tabular}{ll}
\toprule
\multicolumn{2}{l}{\textbf{Input:} Time series $\mathbf{X} = \{x_1, x_2, \ldots, x_T\} \in \mathbb{R}^{T \times 1}$} \\
\multicolumn{2}{l}{\textbf{Output:} Trend value $\beta \in (0,1)$} \\
\midrule
1: & Decompose $\mathbf{X}$ into seasonal, trend, and residual components using STL:
$
\mathbf{X} = \mathbf{S} + \mathbf{T} + \mathbf{R}.
$ \\

2: & Compute the variance of the residual component $\mathbf{R}$. \\

3: & Compute the variance of the combined non-seasonal component $\mathbf{T} + \mathbf{R}$. \\

4: & \textbf{return}
$
\beta = \max\!\left(0,\; 1 - \frac{\mathrm{var}(\mathbf{R})}{\mathrm{var}(\mathbf{T} + \mathbf{R})}\right).
$ \\
\bottomrule
\end{tabular}
}
\end{table}

\clearpage

%% file: Section/APP_re.tex
\section{More Results}
\subsection{Single- and Cross-Dataset Evaluation Results}
\label{app_Single- and Cross-Dataset Evaluation Results}
We report the MAE and MSE results under both single-dataset and cross-dataset learning strategies, evaluated with two alignment schemes across four forecasting horizons \{96, 192, 336, and 720\}, as summarized in Tab.~\ref{tab:mae_results} and Tab.~\ref{tab:mse_results}.

\begin{figure*}[h]
\centering
\includegraphics[width=0.99\textwidth]{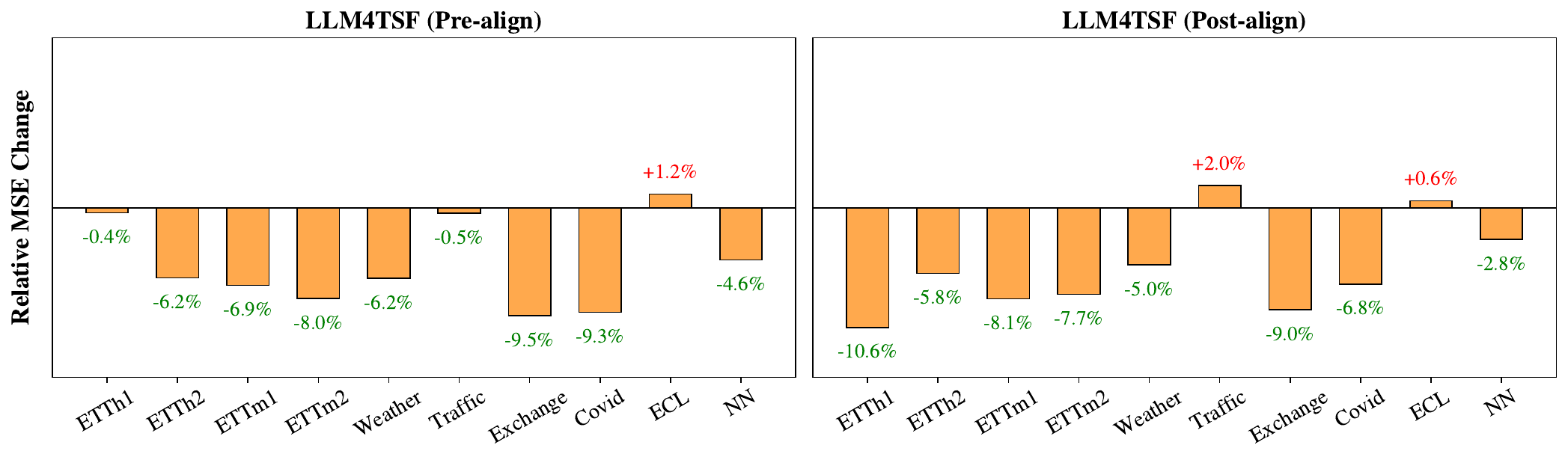}
\caption{Comparison of LLM4TSF performance with pre- and post-alignment under single- and cross-dataset paradigm. 
{\color{green_negative}{Negative}} and \textcolor{red}{Positive} values indicate MSE decreases and increases under cross-dataset learning compared to single-dataset learning.}
\label{fig:llm4tsf_mse}
\end{figure*}

\subsection{Effect of Training Data in Cross-Dataset Learning}
\begin{figure*}[h]
\centering
\includegraphics[width=0.99\textwidth]{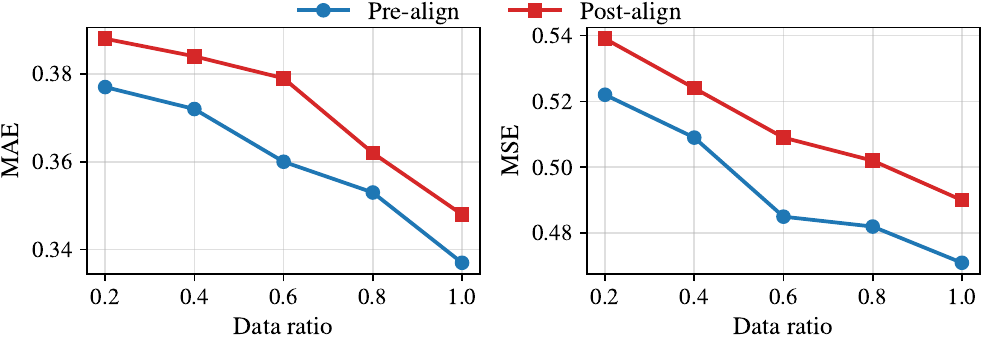}
\caption{Effect of training data on cross-dataset learning performance. Both MAE and MSE decrease as the ratio of data increases.}
\label{fig:data_ratio_mae_mse}
\end{figure*}

During cross-dataset learning, we vary the ratio of training data and gradually increase the available data volume. We observe that both MSE and MAE consistently decrease as the training data ratio increases, indicating a strong correlation between model performance and the amount of training data. This trend suggests that cross-dataset forecasting benefits substantially from larger and more diverse training samples. Results are shown in Fig.~\ref{fig:data_ratio_mae_mse}.

\begin{table*}[t]
\centering
\caption{MAE comparison under different settings.}
\label{tab:mae_results}
\resizebox{0.65\linewidth}{!}{
\begin{tabular}{llcccc}
\toprule
\textbf{Dataset} & \textbf{Horizon} &
\textbf{Pre-Single} & \textbf{Pre-Cross} &
\textbf{Post-Single} & \textbf{Post-Cross} \\
\midrule
ETTh1 & 96  & 0.407 & 0.414 & 0.421 & 0.398 \\
      & 192 & 0.435 & 0.437 & 0.432 & 0.420 \\
      & 336 & 0.452 & 0.448 & 0.462 & 0.446 \\
      & 720 & 0.461 & 0.479 & 0.499 & 0.481 \\
\midrule
ETTh2 & 96  & 0.368 & 0.333 & 0.374 & 0.352 \\
      & 192 & 0.399 & 0.374 & 0.416 & 0.379 \\
      & 336 & 0.435 & 0.411 & 0.446 & 0.418 \\
      & 720 & 0.459 & 0.435 & 0.458 & 0.435 \\
\midrule
ETTm1 & 96  & 0.342 & 0.355 & 0.374 & 0.345 \\
      & 192 & 0.391 & 0.359 & 0.399 & 0.370 \\
      & 336 & 0.417 & 0.377 & 0.428 & 0.396 \\
      & 720 & 0.438 & 0.410 & 0.467 & 0.426 \\
\midrule
ETTm2 & 96  & 0.253 & 0.248 & 0.264 & 0.260 \\
      & 192 & 0.297 & 0.277 & 0.313 & 0.300 \\
      & 336 & 0.347 & 0.328 & 0.369 & 0.344 \\
      & 720 & 0.392 & 0.382 & 0.402 & 0.396 \\
\midrule
Weather & 96  & 0.226 & 0.197 & 0.221 & 0.203 \\
        & 192 & 0.264 & 0.231 & 0.254 & 0.247 \\
        & 336 & 0.286 & 0.269 & 0.295 & 0.281 \\
        & 720 & 0.359 & 0.325 & 0.348 & 0.331 \\
\midrule
Traffic & 96  & 0.254 & 0.241 & 0.271 & 0.276 \\
        & 192 & 0.274 & 0.282 & 0.279 & 0.290 \\
        & 336 & 0.282 & 0.288 & 0.277 & 0.295 \\
        & 720 & 0.311 & 0.302 & 0.300 & 0.310 \\
\midrule
Exchange & 96  & 0.225 & 0.213 & 0.231 & 0.202 \\
         & 192 & 0.312 & 0.294 & 0.325 & 0.296 \\
         & 336 & 0.415 & 0.393 & 0.432 & 0.410 \\
         & 720 & 0.686 & 0.655 & 0.718 & 0.698 \\
\midrule
Covid & 96  & 0.044 & 0.041 & 0.046 & 0.041 \\
      & 192 & 0.049 & 0.045 & 0.052 & 0.048 \\
      & 336 & 0.058 & 0.051 & 0.060 & 0.056 \\
      & 720 & 0.061 & 0.057 & 0.064 & 0.059 \\
\midrule
ECL & 96  & 0.233 & 0.228 & 0.237 & 0.239 \\
    & 192 & 0.259 & 0.253 & 0.251 & 0.254 \\
    & 336 & 0.274 & 0.280 & 0.267 & 0.269 \\
    & 720 & 0.303 & 0.311 & 0.298 & 0.298 \\
\midrule
NN & 96  & 0.613 & 0.584 & 0.641 & 0.622 \\
   & 192 & 0.622 & 0.601 & 0.650 & 0.642 \\
   & 336 & 0.640 & 0.623 & 0.677 & 0.663 \\
   & 720 & 0.691 & 0.659 & 0.723 & 0.709 \\
\bottomrule
\end{tabular}
}
\end{table*}

\begin{table*}[h]
\centering
\caption{MSE comparison under different settings.}
\label{tab:mse_results}
\resizebox{0.65\linewidth}{!}{
\begin{tabular}{llcccc}
\toprule
\textbf{Dataset} & \textbf{Horizon} &
\textbf{Pre-Single} & \textbf{Pre-Cross} &
\textbf{Post-Single} & \textbf{Post-Cross} \\
\midrule
ETTh1 & 96  & 0.404 & 0.412 & 0.420 & 0.377 \\
      & 192 & 0.435 & 0.439 & 0.455 & 0.416 \\
      & 336 & 0.472 & 0.455 & 0.488 & 0.444 \\
      & 720 & 0.485 & 0.481 & 0.566 & 0.488 \\
\midrule
ETTh2 & 96  & 0.303 & 0.279 & 0.307 & 0.283 \\
      & 192 & 0.361 & 0.347 & 0.375 & 0.354 \\
      & 336 & 0.396 & 0.372 & 0.409 & 0.389 \\
      & 720 & 0.425 & 0.395 & 0.425 & 0.403 \\
\midrule
ETTm1 & 96  & 0.322 & 0.291 & 0.336 & 0.290 \\
      & 192 & 0.355 & 0.332 & 0.363 & 0.331 \\
      & 336 & 0.386 & 0.361 & 0.392 & 0.369 \\
      & 720 & 0.454 & 0.426 & 0.449 & 0.427 \\
\midrule
ETTm2 & 96  & 0.171 & 0.165 & 0.179 & 0.169 \\
      & 192 & 0.244 & 0.231 & 0.254 & 0.226 \\
      & 336 & 0.310 & 0.274 & 0.327 & 0.288 \\
      & 720 & 0.371 & 0.339 & 0.388 & 0.377 \\
\midrule
Weather & 96  & 0.158 & 0.151 & 0.161 & 0.147 \\
        & 192 & 0.209 & 0.193 & 0.195 & 0.185 \\
        & 336 & 0.266 & 0.244 & 0.262 & 0.248 \\
        & 720 & 0.325 & 0.311 & 0.335 & 0.322 \\
\midrule
Traffic & 96  & 0.376 & 0.359 & 0.385 & 0.391 \\
        & 192 & 0.391 & 0.403 & 0.404 & 0.413 \\
        & 336 & 0.401 & 0.412 & 0.406 & 0.418 \\
        & 720 & 0.443 & 0.429 & 0.443 & 0.451 \\
\midrule
Exchange & 96  & 0.084 & 0.080 & 0.099 & 0.085 \\
         & 192 & 0.185 & 0.172 & 0.201 & 0.156 \\
         & 336 & 0.368 & 0.321 & 0.379 & 0.327 \\
         & 720 & 0.829 & 0.754 & 1.010 & 0.969 \\
\midrule
Covid & 96  & 1.032 & 1.011 & 1.057 & 1.023 \\
      & 192 & 1.355 & 1.246 & 1.343 & 1.215 \\
      & 336 & 1.689 & 1.318 & 1.657 & 1.518 \\
      & 720 & 2.021 & 1.955 & 2.103 & 1.986 \\
\midrule
ECL & 96  & 0.135 & 0.129 & 0.139 & 0.139 \\
    & 192 & 0.148 & 0.142 & 0.155 & 0.156 \\
    & 336 & 0.158 & 0.166 & 0.170 & 0.171 \\
    & 720 & 0.221 & 0.234 & 0.207 & 0.208 \\
\midrule
NN & 96  & 0.807 & 0.754 & 0.816 & 0.811 \\
   & 192 & 0.824 & 0.786 & 0.833 & 0.829 \\
   & 336 & 0.818 & 0.799 & 0.894 & 0.854 \\
   & 720 & 0.921 & 0.877 & 1.016 & 0.965 \\
\bottomrule
\end{tabular}
}
\end{table*}

\clearpage 
\subsection{Impact of Model Completeness}
\label{app_Impact of Model Completeness}
\begin{figure*}[h]
\centering
\includegraphics[width=0.8\textwidth]{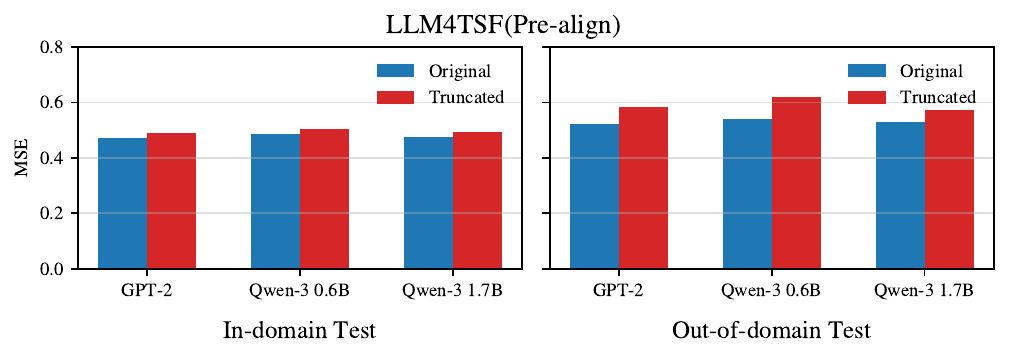}
\includegraphics[width=0.8\textwidth]{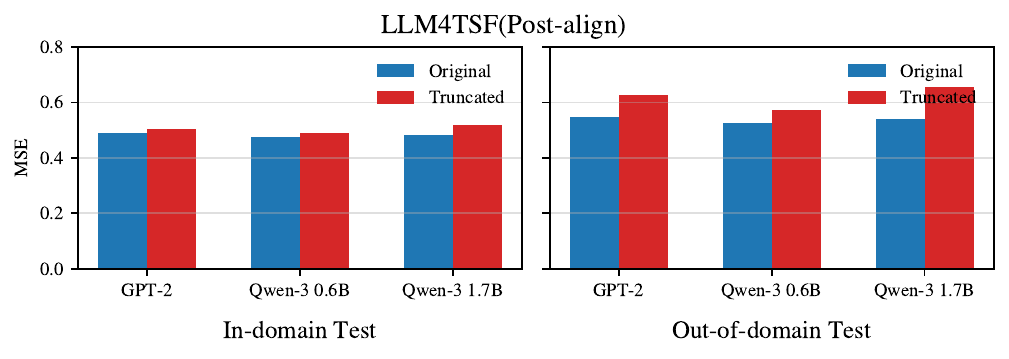}
\caption{Effect of model truncation on forecasting performance.}
\label{fig:truncation}
\end{figure*}

To examine the effect of model completeness, we compare a full LLM with a truncated variant that retains only the first 50\% of layers. Both models are trained on the same data under identical settings and evaluated on in-domain and out-of-domain test sets, as shown in the Fig.~\ref{fig:truncation}. The results indicate that truncating the model degrades performance, leading to higher MSE across evaluation scenarios.

\subsection{Effect of LLM Backbone and Prompt}
\begin{table}[h]
\centering
\caption{Comparison of average MSE and token ratios across different LLM backbones under w/ prompt or w/o prompt.}
\renewcommand{\arraystretch}{1.00}
\resizebox{0.6\linewidth}{!}{
\begin{tabular}{c||cc|cc||cc|cc}
\toprule
\multirow{2}{*}{\textbf{}} 
&\multicolumn{4}{c||}{\textbf{In-Domain Test}} 
& \multicolumn{4}{c}{\textbf{Out-of-Domain Test}} \\ 
&\multicolumn{2}{c}{\textbf{w/ Prompt}} & \multicolumn{2}{c||}{\textbf{w/o Prompt}} & \multicolumn{2}{c}{\textbf{w/ Prompt}} & \multicolumn{2}{c}{\textbf{w/o Prompt}} \\
\midrule
LLM&MSE & Ratio(\%) &MSE & Ratio(\%) &MSE & Ratio(\%) &MSE & Ratio(\%) \\
\midrule
\rowcolor{gray!35}\multicolumn{9}{c}{\textbf{LLM4TSF(Pre-align)}} \\
\midrule
GPT-2 & 0.471 & 46.7 & 0.488\textcolor{myred}{$\uparrow$} & 41.5\textcolor{mygreen}{$\downarrow$} & 0.524 & 72.6 & 0.673\textcolor{myred}{$\uparrow$} & 37.9\textcolor{mygreen}{$\downarrow$} \\ \midrule
\makecell{Qwen-3\\ 0.6B} & 0.485 & 48.2 & 0.495\textcolor{myred}{$\uparrow$} &42.2\textcolor{mygreen}{$\downarrow$} &0.541 & 70.9 &  0.726\textcolor{myred}{$\uparrow$} & 40.6\textcolor{mygreen}{$\downarrow$} \\ \midrule
\makecell{Qwen-3\\ 1.7B} &0.477 & 45.9 & 0.502\textcolor{myred}{$\uparrow$} & 42.9\textcolor{mygreen}{$\downarrow$} & 0.529 & 74.6 & 0.679\textcolor{myred}{$\uparrow$} & 35.5\textcolor{mygreen}{$\downarrow$} \\ \midrule
\rowcolor{gray!35}\multicolumn{9}{c}{\textbf{LLM4TSF(Post-align)}} \\
\midrule
GPT-2 & 0.491 & 53.3&  0.511\textcolor{myred}{$\uparrow$} & 47.7\textcolor{mygreen}{$\downarrow$} & 0.548 & 76.6 & 0.626\textcolor{myred}{$\uparrow$} & 38.5\textcolor{mygreen}{$\downarrow$} \\ \midrule
\makecell{Qwen-3\\ 0.6B} & 0.477 & 55.6 & 0.527\textcolor{myred}{$\uparrow$} & 51.3\textcolor{mygreen}{$\downarrow$} & 0.525 & 73.9 & 0.619\textcolor{myred}{$\uparrow$} & 34.9\textcolor{mygreen}{$\downarrow$}\\ \midrule
\makecell{Qwen-3\\ 1.7B} &0.484 & 52.7 & 0.535\textcolor{myred}{$\uparrow$} & 48.6\textcolor{mygreen}{$\downarrow$} &0.541 & 77.3 & 0.685\textcolor{myred}{$\uparrow$} & 28.8\textcolor{mygreen}{$\downarrow$} \\
\bottomrule 

\end{tabular}}

\label{tab:size}
\vspace{-0.2cm}
\end{table}

\subsection{Baseline Results in Out-of-Domain Tests}
\label{app_Baseline Results in Out-of-Domain Tests}
To assess the out-of-domain generalization of the two alignment strategies after cross-dataset training, we evaluate LLM4TSF (Pre-align) and LLM4TSF (Post-align) on seven unseen datasets, and compare them with three large-scale TS foundation models trained from scratch, namely Chronos~\cite{ansari2024chronos}, UniTS~\cite{gao2024units}, and Moirai~\cite{moirai}. All methods are evaluated under a zero-shot setting, with model configurations and hyperparameters adopted from the original implementations. We further include two LLM-based TSF models, UniTime~\cite{liu2024unitime} and TimeLLM~\cite{jin2023time}, which are trained using single-dataset few-shot learning with only 5\% of the training data, as additional baselines, as shown in Tab.~\ref{tab:mae_baseline_ood} \& \ref{tab:mse_baseline_ood}.
\begin{table*}[h]
\centering
\caption{MAE of baseline out-of-domain test performance.}
\label{tab:mae_baseline_ood}
\resizebox{0.75\linewidth}{!}{
\begin{tabular}{llccccc}
\toprule
\textbf{Dataset} & \textbf{Horizon} &
\textbf{Chronos} & \textbf{UniTS} & \textbf{MOIRAI} & \textbf{UniTime} & \textbf{TimeLLM} \\
\midrule
Wind & 96  & 0.696 & 0.755 & 0.640 & 0.685 & 0.664 \\
     & 192 & 0.767 & 0.823 & 0.722 & 0.786 & 0.767 \\
     & 336 & 0.848 & 0.881 & 0.809 & 0.878 & 0.862 \\
     & 720 & 0.934 & 0.945 & 0.866 & 0.953 & 0.945 \\
\midrule
Solar & 96  & 0.327 & 0.611 & 0.477 & 0.281 & 0.313 \\
      & 192 & 0.339 & 0.655 & 0.526 & 0.270 & 0.377 \\
      & 336 & 0.342 & 0.679 & 0.554 & 0.284 & 0.395 \\
      & 720 & 0.358 & 0.753 & 0.597 & 0.278 & 0.411 \\
\midrule
AQShunyi & 96  & 0.491 & 0.509 & 0.460 & 0.546 & 0.530 \\
         & 192 & 0.522 & 0.538 & 0.475 & 0.578 & 0.551 \\
         & 336 & 0.537 & 0.577 & 0.497 & 0.572 & 0.565 \\
         & 720 & 0.565 & 0.594 & 0.563 & 0.592 & 0.584 \\
\midrule
CzenLan & 96  & 0.246 & 0.487 & 0.477 & 0.377 & 0.318 \\
        & 192 & 0.277 & 0.523 & 0.515 & 0.397 & 0.346 \\
        & 336 & 0.312 & 0.566 & 0.573 & 0.403 & 0.354 \\
        & 720 & 0.388 & 0.638 & 0.622 & 0.434 & 0.399 \\
\midrule
ZafNoo & 96  & 0.399 & 0.553 & 0.404 & 0.603 & 0.485 \\
       & 192 & 0.437 & 0.571 & 0.452 & 0.661 & 0.479 \\
       & 336 & 0.478 & 0.604 & 0.481 & 0.691 & 0.565 \\
       & 720 & 0.507 & 0.669 & 0.514 & 0.724 & 0.577 \\
\midrule
NASDAQ & 96  & 0.494 & 0.769 & 0.577 & 0.614 & 0.503 \\
       & 192 & 0.623 & 0.811 & 0.653 & 0.668 & 0.638 \\
       & 336 & 0.744 & 0.853 & 0.786 & 0.719 & 0.727 \\
       & 720 & 0.786 & 0.860 & 0.813 & 0.885 & 0.822 \\
\midrule
PEMS & 96  & 0.424 & 0.839 & 0.255 & 0.356 & 0.339 \\
     & 192 & 0.486 & 0.854 & 0.271 & 0.364 & 0.375 \\
     & 336 & 0.501 & 0.889 & 0.288 & 0.389 & 0.382 \\
     & 720 & 0.597 & 0.916 & 0.304 & 0.418 & 0.419 \\
\bottomrule
\end{tabular}
}
\end{table*}

\begin{table*}[h]
\centering
\caption{MSE of baseline out-of-domain test performance.}
\label{tab:mse_baseline_ood}
\resizebox{0.75\linewidth}{!}{
\begin{tabular}{llccccc}
\toprule
\textbf{Dataset} & \textbf{Horizon} &
\textbf{Chronos} & \textbf{UniTS} & \textbf{MOIRAI} & \textbf{UniTime} & \textbf{TimeLLM} \\
\midrule
Wind & 96  & 1.250 & 1.038 & 0.963 & 1.022 & 0.981 \\
     & 192 & 1.357 & 1.224 & 1.199 & 1.241 & 1.201 \\
     & 336 & 1.428 & 1.516 & 1.268 & 1.482 & 1.444 \\
     & 720 & 1.653 & 1.653 & 1.513 & 1.688 & 1.658 \\
\midrule
Solar & 96  & 0.418 & 0.779 & 0.858 & 0.216 & 0.402 \\
      & 192 & 0.403 & 0.822 & 0.913 & 0.209 & 0.518 \\
      & 336 & 0.425 & 0.913 & 0.977 & 0.228 & 0.655 \\
      & 720 & 0.488 & 0.968 & 0.995 & 0.220 & 0.733 \\
\midrule
AQShunyi & 96  & 0.733 & 0.855 & 0.607 & 0.868 & 0.788 \\
         & 192 & 0.779 & 0.874 & 0.622 & 0.912 & 0.853 \\
         & 336 & 0.850 & 0.902 & 0.685 & 0.893 & 0.876 \\
         & 720 & 0.871 & 0.928 & 0.759 & 0.945 & 0.920 \\
\midrule
CzenLan & 96  & 0.249 & 0.649 & 0.629 & 0.359 & 0.263 \\
        & 192 & 0.271 & 0.711 & 0.644 & 0.396 & 0.308 \\
        & 336 & 0.308 & 0.768 & 0.671 & 0.399 & 0.318 \\
        & 720 & 0.363 & 0.825 & 0.695 & 0.450 & 0.385 \\
\midrule
ZafNoo & 96  & 0.475 & 0.585 & 0.455 & 0.679 & 0.536 \\
       & 192 & 0.511 & 0.655 & 0.516 & 0.790 & 0.530 \\
       & 336 & 0.560 & 0.694 & 0.579 & 0.853 & 0.645 \\
       & 720 & 0.654 & 0.736 & 0.623 & 0.891 & 0.663 \\
\midrule
NASDAQ & 96  & 0.506 & 0.954 & 0.714 & 0.833 & 0.655 \\
       & 192 & 0.569 & 1.036 & 0.968 & 1.016 & 0.926 \\
       & 336 & 1.112 & 1.255 & 1.253 & 1.188 & 1.033 \\
       & 720 & 1.305 & 1.236 & 1.333 & 1.452 & 1.317 \\
\midrule
PEMS & 96  & 0.485 & 1.094 & 0.159 & 0.349 & 0.337 \\
     & 192 & 0.622 & 1.186 & 0.197 & 0.413 & 0.425 \\
     & 336 & 0.765 & 1.455 & 0.251 & 0.453 & 0.444 \\
     & 720 & 0.872 & 1.476 & 0.365 & 0.461 & 0.456 \\
\bottomrule
\end{tabular}
}
\end{table*}

\clearpage

\subsection{Impact of Statistical Properties}
\label{app_Impact of Statistical Properties}
\begin{figure*}[h]
\centering
\includegraphics[width=0.99\textwidth]{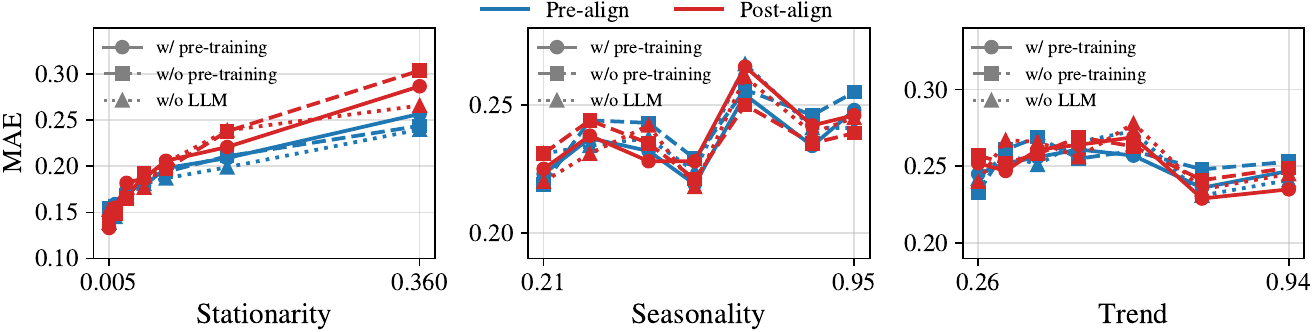}
\caption{Impact of stationarity, seasonality, and trend on forecasting performance}
\label{fig:effect_stationarity_trend}
\end{figure*}

In addition to shifting and transition, we further analyze how variations in stationarity, seasonality, and trend affect the performance of different models. We observe that stationarity plays a non-negligible role in forecasting difficulty: as the degree of stationarity decreases, prediction errors consistently increase. However, under varying stationarity levels, the performance gap between models with and without pre-training, as well as those with and without LLM components, remains relatively small, indicating no significant advantage for either strategy in this regime.
Moreover, our results suggest that changes in seasonality strength and trend magnitude do not fundamentally alter the overall forecasting difficulty, and the relative performance of different models remains largely stable across these settings, as shown in Fig.~\ref{fig:effect_stationarity_trend}.

\begin{figure*}[h]
\centering
\includegraphics[width=0.91\textwidth]{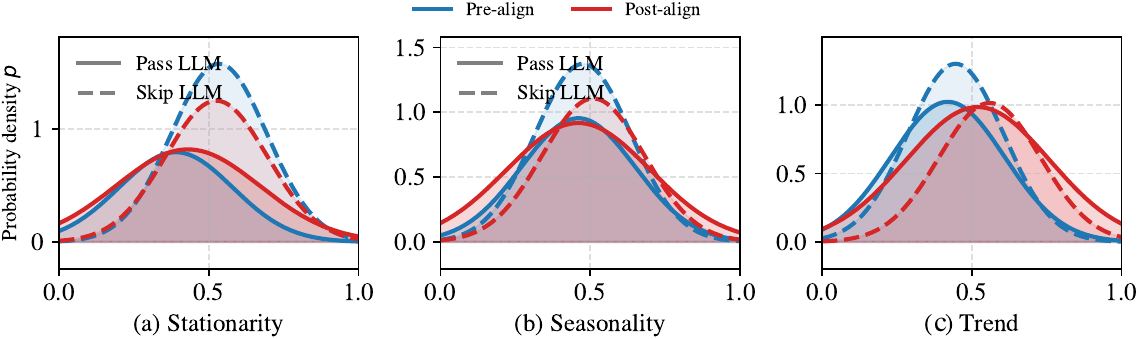}
\caption{When using LLMs w/ pre-training, the distribution of samples passed the LLM is  insensitive to variations in stationarity, seasonality, and trend, exhibiting a relatively uniform pattern.}
\label{fig:distribution_SST}
\end{figure*}

\begin{figure*}[h]
\centering
\includegraphics[width=0.91\textwidth]{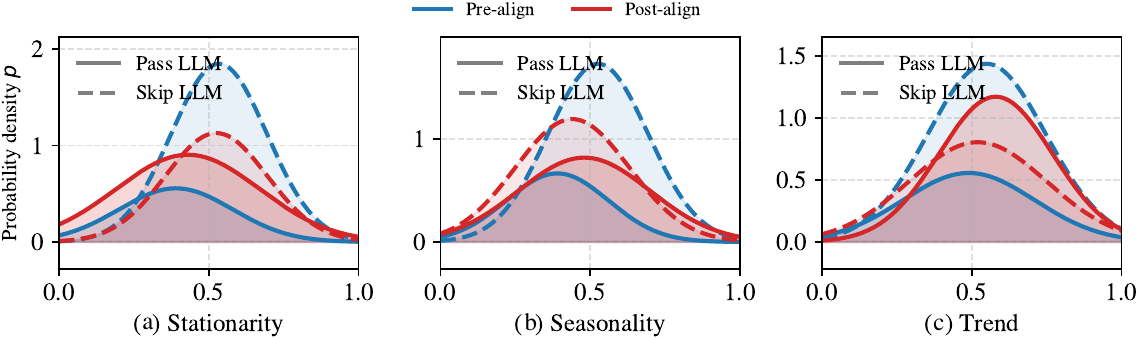}
\caption{Under the LLM w/ pre-training setting, the pass distribution remains uniform with respect to these properties; however, due to the randomly initialized LLM under the pre-alignment scheme, the proportion of skipped samples is higher than that of passed samples.}
\label{fig:distribution_SST1}
\end{figure*}
\clearpage

%% file: Section/APP_sy.tex
\section{Synthetic TS Generation}
\label{app_Synthetic TS Generation}
To isolate the effect of individual temporal properties, we independently construct synthetic TS using property-specific generative operators.
Each operator controls a particular temporal characteristic while preserving intrinsic structure, enabling systematic analysis of forecasting difficulty. For synthetic data generation, we construct 100 time series for each attribute, with each series containing 20,000 observations. We adopt a standard dataset splitting protocol for training, validation, and testing. The forecasting setup uses an input length of 512 time steps to predict the next 192 horizons.
To ensure that the effect of each attribute is examined in isolation and to avoid confounding factors, we strictly separate the synthetic datasets corresponding to different attributes. As a result, time series generated under different attribute controls do not overlap between training and testing phases, preventing interference across attributes during model evaluation.

\textbf{Shifting Synthesis.}\quad
Shifting describes gradual and continuous distributional drift over time.
We model shifting as a time-dependent transformation applied to an underlying latent temporal structure~\cite{he2023domain}.
Let $v(t)$ denote a latent oscillatory process.
A shifted TS is defined as
$
x_t = \mathcal{S}_s\!\big(v\big)(t),
$
where $\mathcal{S}_s(\cdot)$ denotes a shifting operator parameterized by strength $s$.
The operator induces smooth temporal drift through joint modulation of phase, amplitude, and noise statistics, causing the marginal distribution of $x_t$ to vary gradually with time.
Larger values of $s$ correspond to stronger and more complex shifting behavior.

\textbf{Stationarity Synthesis.}\quad
Stationarity describes whether the statistical properties of a TS remain invariant over time.
We synthesize sequences with varying degrees of stationarity by applying time-dependent distributional transformations~\cite{liu2022non}.
Let $v(t)$ be a stationary latent process.
The generated TS is defined as
$
x_t = \sigma_s(t)\,v(t) + \mu_s(t),
$
where $\mu_s(t)$ and $\sigma_s(t)$ denote time-varying mean and scale functions.
Increasing $s$ induces progressively stronger departures from stationarity, while preserving local temporal regularities in $v(t)$.

\textbf{Transition Synthesis.}\quad
Transition characterizes the complexity of temporal dependency structures governing state evolution.
We construct transition-dominated sequences by composing the TS from latent regimes with structured switching behavior~\cite{painblanc2023match}.
Let $\{\pi_k\}_{k=1}^{K}$ denote a finite set of latent temporal patterns.
The generated sequence is given by
$
x_t = \pi_{z_t}(t),
$
where $z_t$ is a latent regime index evolving according to a transition operator
$
z_t \sim \mathcal{T}_s\!\big(z_{1:t-1}\big).
$
Here, $\mathcal{T}_s(\cdot)$ denotes a transition mechanism whose effective dependency order increases with $s$.
As $s$ grows, transitions become increasingly context-dependent, requiring longer temporal history to infer future regimes.

\textbf{Seasonality Synthesis.}\quad
Seasonality reflects recurring temporal patterns across one or multiple time scales.
We generate seasonal TS by modulating periodic structures with time-varying amplitude and phase~\cite{heidrich2023controlling}.
Let $\mathcal{P}(t)$ denote a collection of periodic basis functions.
A seasonal sequence is defined as
$
x_t = \mathcal{A}_s(t)\,\mathcal{P}\!\big(t + \phi_s(t)\big),
$
where $\mathcal{A}_s(t)$ and $\phi_s(t)$ denote amplitude and phase modulation functions, respectively.
The parameter $s$ controls the strength and complexity of seasonal variation, allowing a smooth transition from simple stationary periodicity to multi-scale and non-aligned seasonal patterns.

\textbf{Trend Synthesis.}\quad
Trend captures long-term directional movement that is not necessarily governed by a globally extrapolatable function~\cite{lin2021ssdnet}.
We construct trend-dominated sequences by composing bounded temporal patterns with structured trend components.
Let $v(t)$ denote a bounded latent pattern.
A trend-augmented TS is given by
$
x_t = v(t) + g_s(t),
$
where $g_s(t)$ denotes a trend function whose functional form and smoothness vary with $s$.
Larger values of $s$ correspond to more complex and non-linear trend behavior, yielding increasing difficulty for parametric extrapolation methods.

\begin{figure*}[h]
\centering
\includegraphics[width=0.67\textwidth]{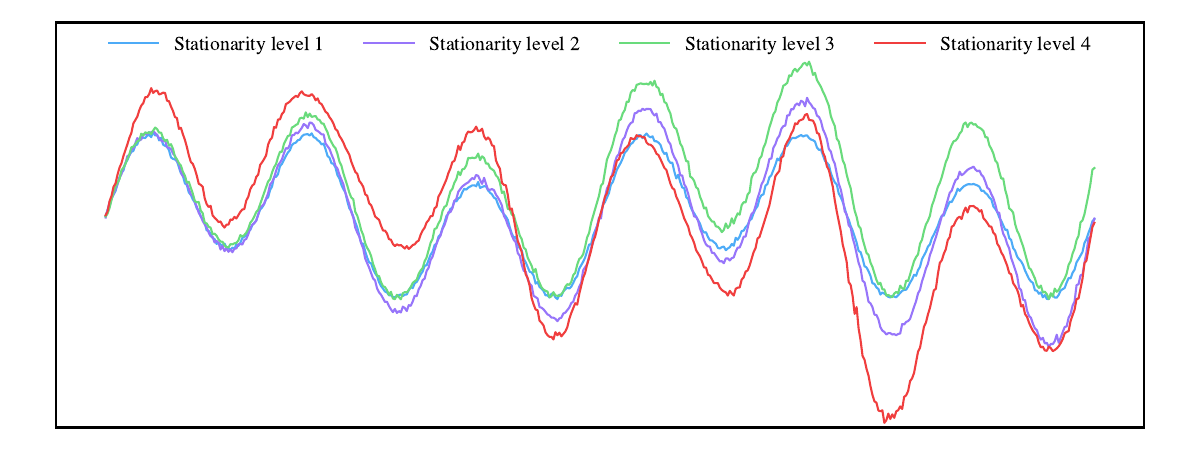}
\includegraphics[width=0.67\textwidth]{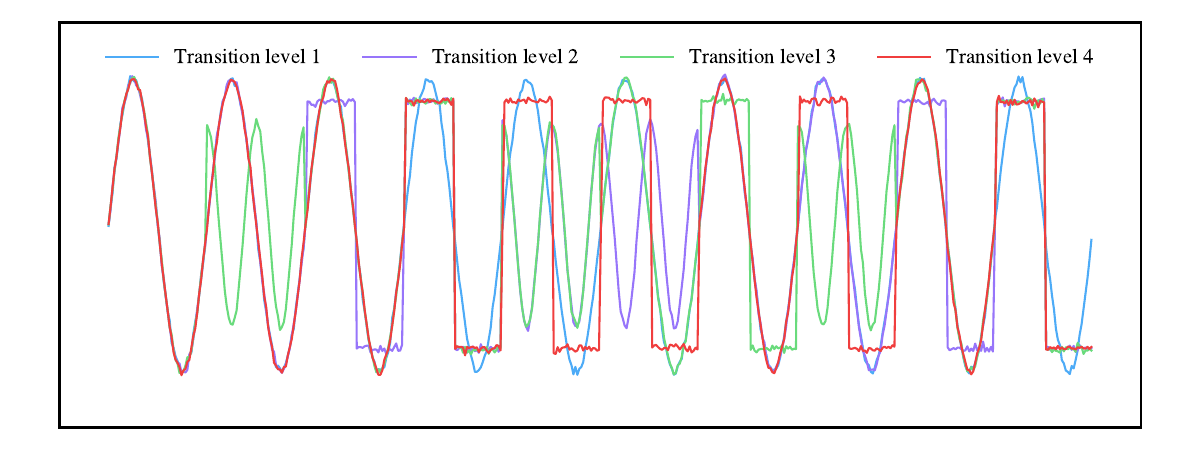}
\includegraphics[width=0.67\textwidth]{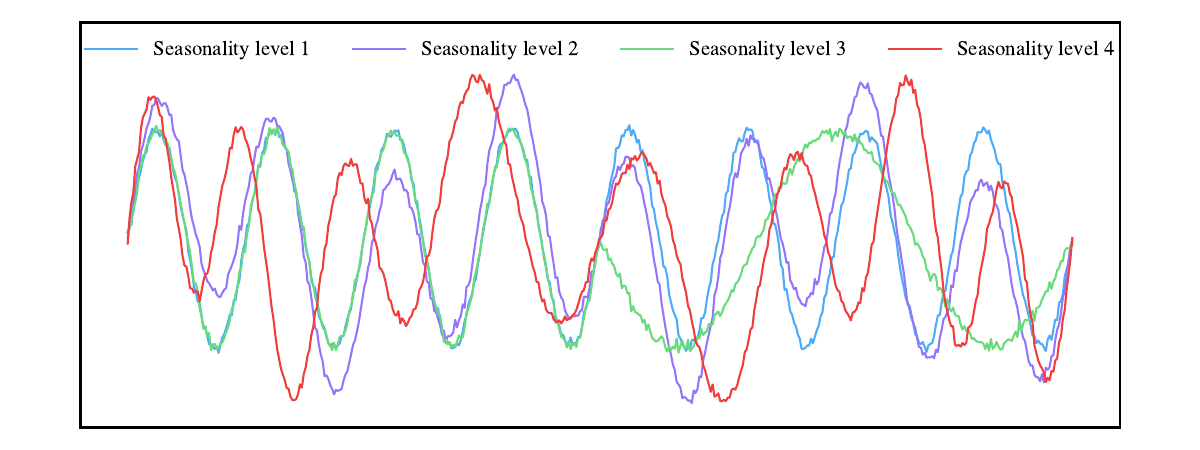}
\includegraphics[width=0.67\textwidth]{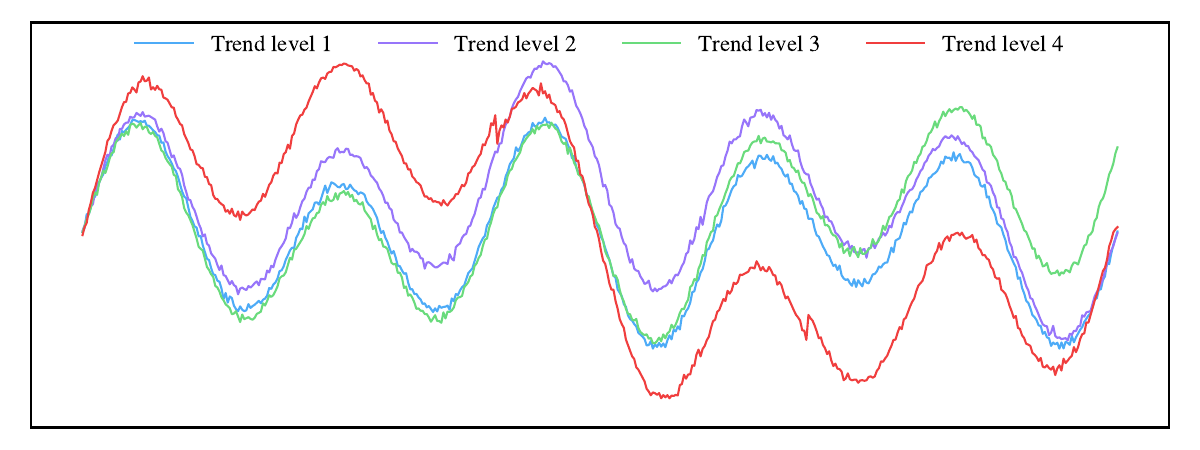}

\caption{Illustration of synthetic TS generated under different property control levels, where higher levels indicate stronger effects.}
\label{fig:synthetic TS1}
\end{figure*}

Across all properties, the strength parameter $s$ controls temporal complexity while preserving underlying structure.
This design ensures that forecasting difficulty increases in a controlled and interpretable manner, favoring models capable of capturing long-range dependencies and adaptive temporal representations. Five types of synthetic TS generated by controlling different statistical properties, as illustrated in the Fig.~\ref{fig:synthetic TS1}.

\begin{figure*}[h]
\centering
\includegraphics[width=0.67\textwidth]{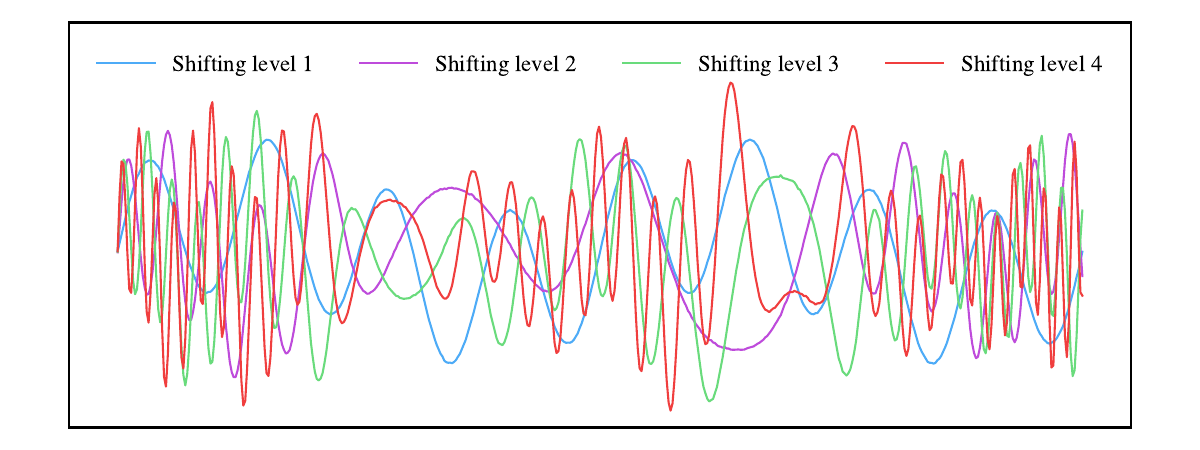}
\label{fig:synthetic TS}
\end{figure*}

\clearpage

%% file: Section/APP_router.tex
\section{Router Mechanism}
\label{app_Router Mechanism}
\subsection{Gumbel-Softmax and STE}
For each token, a routing module generates a pair of logits
$\mathbf{z} = [z_1, z_2]$, representing the routing preferences over two possible paths.
To obtain a discrete routing decision, we employ the Gumbel-Max trick~\cite{jang2017categoricalreparameterizationgumbelsoftmax}, where independent
Gumbel noise variables $g_i \sim \mathrm{Gumbel}(0,1)$ are added to the logits, and the routing outcome is sampled as:
$$
\mathbf{y}_{\text{hard}} =
\text{one-hot}\!\left(
\arg\max_i \left( z_i + g_i \right)
\right).
$$

However, the $\arg\max(\cdot)$ operation is non-differentiable, which prevents gradients from propagating through the routing decision.
To overcome this limitation, we adopt the Gumbel-Softmax relaxation to obtain a continuous and differentiable approximation of the discrete routing variable:
$$
y_i =
\frac{\exp\!\left((z_i + g_i)/\tau\right)}
{\sum_j \exp\!\left((z_j + g_j)/\tau\right)},
$$
where $\tau$ is a temperature parameter that controls the smoothness of the resulting distribution.
As $\tau$ decreases, the output distribution becomes increasingly peaked, approaching a one-hot representation.

In practice, we further integrate the Gumbel-Softmax mechanism with the Straight-Through Estimator (STE)~\cite{bengio2013estimating}.
During the forward pass, the discrete routing decision $\mathbf{y}_{\text{hard}}$ is used to perform token-level path selection,
while in the backward pass, gradients are propagated through the continuous approximation $\mathbf{y}_{\text{soft}}$.
This is achieved by constructing the final routing variable as:
$$
\mathbf{y} =
\mathbf{y}_{\text{hard}}
- \operatorname{detach}(\mathbf{y}_{\text{soft}})
+ \mathbf{y}_{\text{soft}}.
$$
This design enables effective discrete routing at inference time while maintaining differentiability during training.

\subsection{Algorithm: Token-Level Routing}
\begin{table}[H]
\centering
\caption{Algorithm: Token-Level Routing}
\label{tab:routing}
\resizebox{0.80\linewidth}{!}{
\begin{tabular}{ll}
\toprule
\multicolumn{2}{l}{\textbf{Require:} Dataset $D$, routing module parameters $\theta$, temperature $\tau$} \\
\multicolumn{2}{l}{\textbf{Output:} Differentiable token-level routing decisions $\mathbf{y}$} \\
\midrule
1: & Initialize routing module parameters $\theta$ \\

2: & \textbf{for each training iteration do} \\

3: & \quad Sample a token of TS segments $X \sim D$ \\

4: & \quad \textbf{for each token $t$ in $X$ do} \\

5: & \qquad Compute routing logits $\mathbf{z}_t = [z_{t,1}, z_{t,2}]$ using the routing module \\

6: & \qquad Sample Gumbel noise $g_i \sim \mathrm{Gumbel}(0,1)$ for each routing option \\

7: & \qquad Obtain discrete routing decision using Gumbel-Max: \\
   & \qquad $\mathbf{y}_{t}^{\text{hard}} \leftarrow \text{one-hot}\!\left(\arg\max_i (z_{t,i} + g_i)\right)$ \\

8: & \qquad Compute continuous routing weights via Gumbel-Softmax relaxation: \\
   & \qquad $y_{t,i}^{\text{soft}} \leftarrow
   \dfrac{\exp((z_{t,i}+g_i)/\tau)}{\sum_j \exp((z_{t,j}+g_j)/\tau)}$ \\

9: & \qquad Apply Straight-Through Estimator (STE): \\
   & \qquad $\mathbf{y}_t \leftarrow
   \mathbf{y}_{t}^{\text{hard}}
   - \operatorname{detach}(\mathbf{y}_{t}^{\text{soft}})
   + \mathbf{y}_{t}^{\text{soft}}$ \\

10: & \qquad Route token $t$ according to $\mathbf{y}_t$ \\

11: & \quad \textbf{end for} \\

12: & \quad Backpropagate gradients through $\mathbf{y}_{\text{soft}}$ and update $\theta$ \\

13: & \textbf{end for} \\

14: & \textbf{return} learned routing mechanism \\

\bottomrule
\end{tabular}
}
\end{table}

%% file: Section/APP_repro.tex
\section{Reproducibility Details}
To ensure that the results and conclusions can be reproduced accurately, we provide the exact training configurations used in our experiments, as summarized in the Tab.~\ref{tab:training_config}. Additional details of the routing analysis are provided in Tab.~\ref{tab:routing_config}.
\begin{table}[htbp]
\centering
\caption{Training Configuration}
\label{tab:training_config}
\resizebox{0.5\linewidth}{!}{
\begin{tabular}{ll}
\toprule
\textbf{Hyperparameter} & \textbf{Value} \\
\midrule
Framework & HuggingFace Transformers \\
Distributed Training & DeepSpeed, Accelerate \\
GPU & $4\times$ NVIDIA A100 40GB / H100 80GB \\
Random Seed & Fixed 2026 \\
Optimizer & AdamW\\
Learning Rate & $1 \times 10^{-4}$ \\
Adam $\beta_1$ & 0.9 \\
Adam $\beta_2$ & 0.95 \\
Adam $\epsilon$ & $1 \times 10^{-6}$ \\
Weight Decay & 0.01 \\
LR Scheduler & Cosine decay \\
Warmup Ratio & 0 \\
Precision & bf16 \\
DeepSpeed Stage & ZeRO Stage 2 \\
Micro Batch Size per GPU & 64 \\
Gradient Accumulation Steps & 1 \\
Gradient Clipping & Default \\
\bottomrule
\end{tabular}
}
\end{table}

\begin{table}[htbp]
\centering
\caption{Routing Analysis Configuration}
\label{tab:routing_config}
\resizebox{0.48\linewidth}{!}{
\begin{tabular}{ll}
\toprule
\textbf{Routing Parameter} & \textbf{Value} \\
\midrule
Number of Routing Paths & 2 \\
Routing Module Output & Logits $\mathbf{z} = [z_1, z_2]$ \\
Gumbel Noise Distribution & Gumbel$(0,1)$ \\
Temperature $\tau$ (init) & 1.0 \\
Temperature $\tau$ (final) & 0.1 \\
Discrete Sampling & Gumbel-Max for $\mathbf{y}_{\text{hard}}$ \\
Straight-Through Estimator & Enabled \\
Forward Pass Routing & Hard one-hot $\mathbf{y}_{\text{hard}}$ \\
Backward Pass Gradient & Soft $\mathbf{y}_{\text{soft}}$ \\
Inference-time Routing & $\arg\max(\mathbf{z})$ \\
Routing Regularization & Entropy penalty on $\mathbf{y}_{\text{soft}}$ \\
Entropy Weight & $1\times10^{-3}$ \\
Target Routing Ratio & 0.5 \\
Ratio Penalty Weight & $1\times10^{-2}$ \\
Router Optimizer & AdamW \\
Router Learning Rate & $3\times10^{-4}$ \\
Weight Decay & 0.01 \\
Gradient Clipping & 1.0 \\
\bottomrule
\end{tabular}
}
\end{table}

\clearpage